\documentclass[lettersize,journal]{IEEEtran}
\usepackage{amsmath,amsfonts}
\usepackage{algorithmic}
\usepackage{array}
\usepackage[caption=false,font=normalsize,labelfont=sf,textfont=sf]{subfig}
\usepackage{textcomp}
\usepackage{stfloats}
\usepackage{url}
\usepackage{verbatim}
\usepackage{graphicx}
\def\BibTeX{{\rm B\kern-.05em{\sc i\kern-.025em b}\kern-.08em
    T\kern-.1667em\lower.7ex\hbox{E}\kern-.125emX}}
\usepackage{balance}

\usepackage{amsmath} 
\usepackage{amssymb}
\usepackage{url}            % simple URL typesetting
\usepackage{booktabs}       % professional-quality tables
\usepackage{amsfonts}       % blackboard math symbols
\usepackage{nicefrac}       % compact symbols for 1/2, etc.
\usepackage{microtype}      % microtypography
\usepackage{xcolor}         % colors  *******************
\usepackage{array}   % 用于自定义列格式
\usepackage{booktabs} % 提供更好的表格线
\usepackage{graphicx}
\usepackage{amsthm}
\usepackage{multirow}
\usepackage{booktabs}
\usepackage{array}
\usepackage{caption}
\usepackage{url}
\usepackage{xr-hyper}
\usepackage{hyperref}

\begin{document}
% \makeatletter
% \xapptocmd{\NAT@bibsetnum}{\setlength{\leftmargin}{0pt}\setlength{\itemindent}{\labelwidth}\addtolength{\itemindent}{\labelsep}}{}{}
% \makeatother

\title{Accurate Precipitation Forecast by Efficiently Learning from Massive Atmospheric Variables and Unbalanced Distribution}
% All-weather意义不大，因为晴空占比过大，这部分规律只需要让模型习得即可，不必使其数据占比过大

% The \author macro works with any number of authors. There are two commands
% used to separate the names and addresses of multiple authors: \And and \AND.
%
% Using \And between authors leaves it to LaTeX to determine where to break the
% lines. Using \AND forces a line break at that point. So, if LaTeX puts 3 of 4
% authors names on the first line, and the last on the second line, try using
% \AND instead of \And before the third author name.

% \author{%
%   Shuangliang Li\textsuperscript{1,2} \\
%   \texttt{whu\_lsl@whu.edu.cn} \\
%   \and
%   Siwei Li\textsuperscript{1,2}\thanks{Corresponding author: Siwei Li, siwei.li@whu.edu.cn} \\
%   \texttt{siwei.li@whu.edu.cn} \\
%   \and
%   Li Li\textsuperscript{3} \\
%   \texttt{63477223@qq.com} \\
%   \and
%   Weijie Zou\textsuperscript{1} \\
%   \texttt{2016301610343@whu.edu.cn} \\
%   \and
%   Jie Yang\textsuperscript{1,2} \\
%   \texttt{jie.yang@whu.edu.cn}
%   \and
%   Maolin Zhang\textsuperscript{1,2} \\
%   \texttt{maolinzhang@whu.edu.cn}
% }
% \date{}  % 留空

\author{Shuangliang Li, Siwei Li*, Li Li, Weijie Zou, Jie Yang, Maolin Zhang
        % <-this % stops a space
 \thanks{
 This work was supported by the Fundamental Research Funds for the Central Universities
 
 S. Li, S. Li, Weijie Zou, Jie Yang and Maolin Zhang are with the Hubei Key Laboratory of Quantitative Remote Sensing of Land and Atmosphere, School of Remote Sensing and Information Engineering, Wuhan University, Wuhan, 430079, China and Perception and Effectiveness Assessment for Carbon-neutrality Efforts, Engineering Research Center of Ministry of Education, Institute for Carbon Neutrality, Wuhan University, Wuhan, 430072, China (e-mail: whu\_{lsl}@whu.edu.cn; siwei.li@whu.edu.cn; 2016301610343@whu.edu.cn; jie.yang@whu.edu.cn; maolinzhang@whu.edu.cn)\emph{(Corresponding author: Siwei Li.)}

Li Li is with the Hubei Meteorological Information and Technology Support Center, Wuhan, 430074, China (e-mail: 63477223@qq.com)
 }}
% \author{%
%   \begin{minipage}[t]{0.47\linewidth}
%     \centering
%     Shuangliang Li\textsuperscript{1,2}\\                   % 作者 1
%     \texttt{whu\_lsl@whu.edu.cn}\\[6pt]
%     Li Li\textsuperscript{3}\\  % 作者 2
%     \texttt{63477223@qq.com}
%   \end{minipage}\hfill
%   \begin{minipage}[t]{0.47\linewidth}
%     \centering
%     Siwei Li\textsuperscript{1,2}\textsuperscript{*}\\      % 作者 3
%     \texttt{siwei.li@whu.edu.cn}\\[6pt]
%     Weijie Zou\textsuperscript{1}\\                        % 作者 4
%     \texttt{2016301610343@whu.edu.cn}
%   \end{minipage}\\[12pt]
%   \begin{minipage}[t]{0.47\linewidth}
%     \centering
%     Jie Yang\textsuperscript{1,2}\\                        % 作者 5
%     \texttt{jie.yang@whu.edu.cn}
%   \end{minipage}\hfill
%   \begin{minipage}[t]{0.47\linewidth}
%     \centering
%     Maolin Zhang\textsuperscript{1,2}\\                    % 作者 6
%     \texttt{maolinzhang@whu.edu.cn}
%   \end{minipage}%
% }

% \begin{document}
% \linenumbers

\maketitle

% % This work was supported by the Fundamental Research Funds for the Central Universities.
% \footnotetext[1]{Hubei Key Laboratory of Quantitative Remote Sensing of Land and Atmosphere, School of Remote Sensing and Information Engineering, Wuhan University, Wuhan, 430079, China}
% \footnotetext[2]{Perception and Effectiveness Assessment for Carbon-neutrality Efforts, Engineering Research Center of Ministry of Education, Institute for Carbon Neutrality, Wuhan University, Wuhan, 430072, China}
% \footnotetext[3]{Hubei Meteorological Information and Technology Support Center, Wuhan, 430074, China}
% % \footnotetext[4]{This work was supported by the Fundamental Research Funds for the Central Universities.}
% \vspace{1em}
% \begin{center}
% \textsuperscript{1}Hubei Key Laboratory of Quantitative Remote Sensing of Land and Atmosphere, School of Remote Sensing and Information Engineering, Wuhan University, Wuhan, 430079, China\\[4pt]
% \textsuperscript{2}Perception and Effectiveness Assessment for Carbon-neutrality Efforts, Engineering Research Center of Ministry of Education, Institute for Carbon Neutrality, Wuhan University, Wuhan, 430072, China\\[4pt]
% \textsuperscript{3}Hubei Meteorological Information and Technology Support Center, Wuhan, 430074, China\\[8pt]
% \textsuperscript{*}Corresponding author: Siwei Li, siwei.li@whu.edu.cn
% \end{center}

\begin{abstract}
% 当预报时间增加时，降水事件所占比例会降低，可能会漏检，需要增加对漏检的惩罚。
%最近深度学习被越来越多地应用于降水预报。然而，当前大部分研究为超短期临近降水预测（nowcast），而非更有意义和实用的continuous 短中期预报。这主要是由于降水事件的演变规律过于复杂,而降水样本又相对稀少，并且现有的模型难以高效地利用大量的多源大气观测数据(一般只使用雷达数据)，使得现有神经网络很难捕捉到降水的出现和演变规律。针对上述问题，本文提出了整合更多大气观测的隐空间迭代预报模型，通过仅仅预测和降水演变相关的隐特征来高效准确地模拟降水系统的演变。并且针对降水事件的极端长尾分布特性，本研究设计了ELT-CE损失函数来增加对漏检事件的惩罚，并配合像素级loss实现了对降水强度的准确预测。在两个不同时空分辨率的数据集上的定量和定性的比较实验结果充分证明了所提出的预报模型和损失函数的有效性和高效性。
Short-term (0-24 hours) precipitation forecasting is highly valuable to socioeconomic activities and public safety. However, the highly complex evolution patterns of precipitation events, the extreme imbalance between precipitation and non-precipitation samples, and the inability of existing models to efficiently and effectively utilize large volumes of multi-source atmospheric observation data hinder improvements in precipitation forecasting accuracy and computational efficiency.
To address the above challenges, this study developed a novel forecasting model capable of effectively and efficiently utilizing massive atmospheric observations by automatically extracting and iteratively predicting the latent features strongly associated with precipitation evolution. Furthermore, this study introduces a ‘WMCE’ loss function, designed to accurately discriminate extremely scarce precipitation events while precisely predicting their intensity values.
Extensive experiments on two datasets demonstrate that our proposed model substantially and consistently outperforms all prevalent baselines in both accuracy and efficiency. Moreover, the proposed forecasting model substantially lowers the computational cost required to obtain valuable predictions compared to existing approaches, thereby positioning it as a milestone for efficient and practical precipitation forecasting.
\end{abstract}


\section{Introduction}\label{sec1}
% 对比本文构建的降水预测框架和其他方法的预测框架。LSTM/PredRNN主要关注于短期降水预测。Metnet-x利用了基于密集区间划分的交叉熵损失函数来训练单时刻预测网络，而我们的迭代预测框架结合新设计的基于交叉熵的损失函数和像素级loss来训练网络并预测短中期降水。
% \begin{figure*}[ht]
% \centering
%     \includegraphics[width=1\textwidth,trim=0 0 0 0,clip]{picture/general/band1_histogram_with_log.png}
%     \caption{Distribution Analysis: Raw vs. Log-Transformed Near-Surface Radar Reflectivity (\enquote{USA} Dataset). Please note the units of the vertical axis ($1e8$).
% }
%     \label{fig:radar_hist}
% \end{figure*}
% \begin{table}[ht]
% \centering
% \caption{Hourly cumulative Precipitation Rates (mm/h) and Radar Reflectivity (dBZ) Value Distribution by Bucket. Note that the radar reflectivity exceeding 20 dBZ generally corresponds to the presence of precipitation events with the precipitation rate approximately great than 0.2 mm/h \cite{MRMS}\cite{serafin1990meteorological}\cite{wang2025high}.}
\begin{table}[ht]
\centering
\caption{Hourly cumulative Precipitation Rates (mm/h) and Radar Reflectivity (dBZ) Value Distribution by Bucket. Note that the radar reflectivity exceeding 20 dBZ generally corresponds to the presence of precipitation events with the precipitation rate approximately great than 0.2 mm/h \cite{MRMS}\cite{serafin1990meteorological}\cite{wang2025high}.}
\begin{tabular}{l *{6}{r}} % 左对齐第一列，右对齐6列数据列
\toprule
& \multicolumn{6}{c}{\textbf{Cumulative Precipitation Rates (QPE, mm/h)}} \\
\cmidrule(lr){2-7} % 覆盖第2列到第7列
\textbf{Measure Type} & \textbf{$\le$0.2} & \textbf{0.2-1} & \textbf{1-2} & \textbf{2-4} & \textbf{4-8} & \textbf{$\ge$8} \\
\midrule
‘USA’ Dataset & 93.31\% & 3.45\% & 1.42\% & 0.97\% & 0.52\% & 0.33\%  \\
\midrule
& \multicolumn{6}{c}{\textbf{Radar Reflectivity (dBZ)}} \\
\cmidrule(lr){2-7} % 覆盖第2列到第7列
\textbf{Measure Type} & \textbf{$\le$20} & \textbf{20-25} & \textbf{25-30} & \textbf{30-35} & \textbf{35-40} & \textbf{$\ge$40} \\
\midrule
‘Hubei’ Dataset& 98.28\% & 0.82\% & 0.51\% & 0.26\% & 0.10\% & 0.04\% \\
\bottomrule
\end{tabular}
% \caption{请在此处添加表格标题} % 建议添加标题
\label{tab:tab0}
\end{table}
\begin{figure*}[ht]
\centering
    \includegraphics[width=1\textwidth,trim=0 55 0 0,clip]{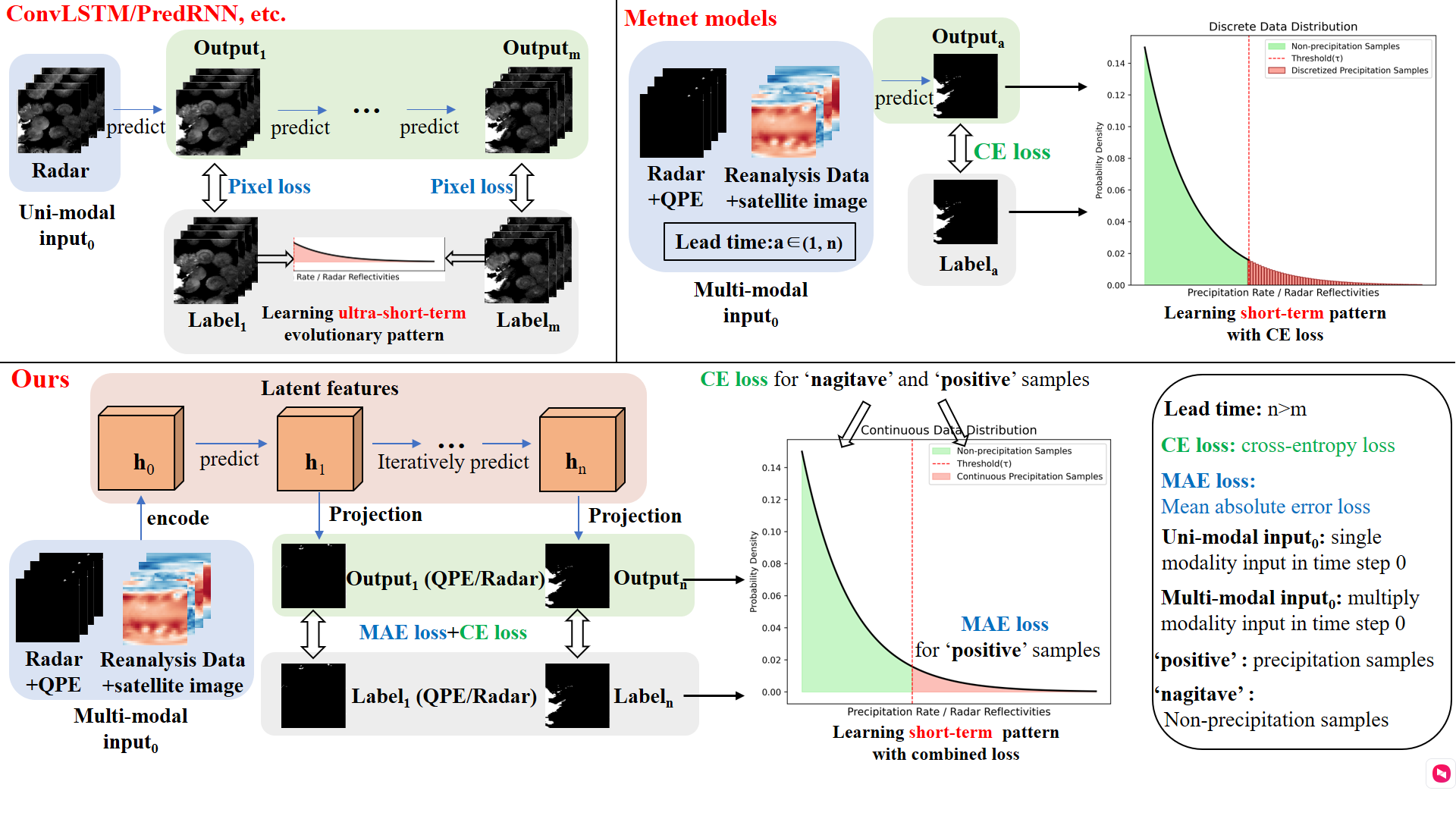}
    \caption{
    Comparison between the precipitation prediction framework. ConvLSTM \cite{NIPS2015_07563a3f} and PredRNN \cite{9749915} primarily focus on ultra-short-term severe convection and thunderstorms nowcasting. Metnet models \cite{snderby2020metnet}\cite{Espeholt2022}\cite{andrychowicz2023} employs a dense interval-partitioned cross-entropy loss to train its single-lead-time network. In contrast, our novelly designed latent space iterative prediction framework integrates a ‘WMCE’ loss to train the network for short-term precipitation forecasting. 
    % Note that our model takes two time steps as input in a single iteration and predicts one time step, while the above diagram presents a simplified representation.
}
    \label{fig:overall_framework}
\end{figure*}
% 降水预报指预测未来一段时间内降水发生时间和演变规律（如果发生）。近地面雷达回波强度和降水强度具有很强的相关性，降水强度可以通过Z-R关系式从回波强度反演得到。如图显示当雷达回波强度大于20时我们认为发生了降水。因此预报雷达回波强度和预报降水强度的演变规律具有等价性，对防灾减灾、农业生产、水资源管理和交通出行都非常重要。本文将预报雷达回波强度，并且在表述上不和降水预报区分。  雨量计数据难以获取/在雨量计稀疏地区影响降水率反演精度。因此我们选择直接预测近地面雷达反射率因子，其和地面的降水率相关性非常强。
Precipitation forecasting refers to predicting the timing and evolution patterns of precipitation (if it occurs) over a future period.
% As shown in Fig. \ref{fig:overall_framework}, precipitation is considered to occur when precipitation rate exceeds 0.2 or radar reflectivity exceeds 20. The radar reflectivity factor (dBZ) exhibits a strong correlation with precipitation intensity \cite{BERNE2004166}. And precipitation rates (mm/hr) can be derived from radar measurements using the Z-R relationship or the Marshall-Palmer formula\footnote{https://web.archive.org/web/20160113151652/http://www.desktopdoppler.com/help/nws-nexrad.htm} or more complex conversion formulas \cite{MRMS}. Therefore, forecasting radar reflectivity is equivalent to forecasting the evolution of precipitation intensity, 
Accurate precipitation forecasting is critically important for disaster prevention and mitigation, agricultural production, water resource management, and transportation \cite{Ravuri2021}\cite{Zhang2023}\cite{Song2023}. In this paper, we forecast both precipitation rate and radar reflectivity, and these terms are used interchangeably with "precipitation forecasting" throughout the text.

Traditional numerical weather prediction (NWP) models simulate the evolution of precipitation by solving a series of partial differential equations. But it requires execution on a supercomputer to obtain the forecast results, which consumes a substantial amount of computational resources and time \cite{Bauer2015}\cite{ScientificChallenges}.

Recently, deep learning methods have been increasingly applied to precipitation prediction due to their fast inference speed and strong capability to learn evolutionary patterns of precipitation events \cite{NIPS2015_07563a3f}\cite{9749915}. However, most contemporary precipitation forecasting models 
% achieve effective high-accuracy predictions primarily for nowcasting (0-2 hour lead times). 
% And they inevitably deliver poor predictive results when applied to short-to-medium-range forecasting scenarios. This is because precipitation exhibits highly complex and diverse patterns of variation over longer time ranges.
% Actually, existing deep learning-based precipitation forecasting models commonly 
suffer from limited accuracy or low efficiency in short-term forecasting, primarily due to the following three reasons:

(1) Most existing studies focus on ultra-short-term nowcasting of severe convection and thunderstorms (with lead times of 0-2 hours) and rely solely on single-source atmospheric observations (primarily weather radar data), as shown in the top left corner of Fig. \ref{fig:overall_framework}. For example, Shi et al. \cite{NIPS2015_07563a3f} introduced a recurrent neural network architecture that utilizes convolutional LSTMs \cite{Graves2012}. Subsequently, Wang et al. developed PredRNN \cite{NIPS2017_e5f6ad6c}, which enhances spatiotemporal modeling with zigzag memory flow and decoupled memory cells. Ravuri et al. \cite{Ravuri2021} developed a deep generative model for probabilistic nowcasting, while Zhang et al. \cite{Zhang2023} proposed NowcastNet to predict extreme precipitation at kilometer scales. However, when the prediction time range is extended to 0-24 hours, predicting occurrence and evolution of precipitation events from single-source data alone is extremely difficult. Because the atmosphere is an extremely complex system and multi-source data are required to accurately simulate its evolution. 

(2) Although some studies have recognized the importance of multi-source atmospheric observations, the manners employed by these models to predict multiple lead time steps remain deficient. For example, Metnet models \cite{snderby2020metnet}\cite{Espeholt2022}\cite{andrychowicz2023} take multi-source data as input and accurately predict the probability of different precipitation intensity events occurring at a specific lead time within the next 0-8/12/24 hours, as shown in the top right corner of Fig. \ref{fig:overall_framework}. In contrast, Ning et al. \cite{ning2023mimo} proposed a Multi-In-Multi-Out (MIMO) architecture that outputs all future frames at once. However, both the ‘single lead-time’ prediction approach and the multi-time-step simultaneous prediction manner entail high task complexity and require massive computational resources to achieve high prediction accuracy and strong temporal correlation. This, in turn, makes tuning hyperparameters and network architectures extremely difficult.

(3) To reduce the complexity of the task of directly outputting a single time step or all time steps, some studies have adopted iterative prediction methods to accomplish multi-time-steps forecasting. Iterative prediction is an efficient way to simulate the evolution of atmospheric systems with the inductive bias of strong temporal correlation. However, most existing deep learning models that iteratively predict atmospheric variables including precipitation intensity operate in the original physical variable space \cite{pathak2022fourcastnet}\cite{doi:10.1126/science.adi2336}. This requires the model to learn the evolution patterns of all input variables at each time step, resulting in significantly increased learning difficulty and unnecessary computational costs since not all available multi-level atmospheric observations are strongly associated with precipitation \cite{Espeholt2022}\cite{li2024efficiently}. Therefore, it is crucial to accurately extract atmospheric features that are strongly associated with precipitation evolution and to perform iterative predictions based on them. However, manually distinguishing precipitation-related atmospheric signals from irrelevant observations is notoriously difficult in practice. Thus, how to accurately and automatically extract precipitation-related atmospheric features is of vital importance.

Moreover, directly applying iterative prediction models to forecast extremely scarce precipitation events remains highly challenging.  
As shown in Table \ref{tab:tab0}, precipitation samples (with precipitation rates exceeding 0.2 mm/h or radar reflectivity factors above 20 dBZ) account for an extremely small proportion, exhibiting a severe imbalance with the large number of non-precipitation samples.  
And existing pixel-level loss functions (such as MSE or MAE) primarily focus on learning high-probability events and perform poorly for rare samples.
Although some strategies, such as ‘importance sampling’ \cite{Espeholt2022}, can increase the proportion of precipitation samples, we experimentally found that such strategies actually degrade performance. This may be because some sequences with very low radar echo intensity could serve as precursor signals for subsequent precipitation, especially in tasks predicting longer lead times. Therefore, a new loss function must be designed to improve prediction accuracy for rare precipitation samples. 

To solve above problems, this study make several contributions as follows:  

1. To efficiently and effectively leverage large volumes of atmospheric observation data, we introduce an iterative prediction framework operating in a encoded low-dimensional latent space (as illustrated in Figs. \ref{fig:overall_framework} and \ref{fig:network}). The latent representations are optimized through loss functions to be strongly associated with precipitation initiation and evolution. 
% 1. To reduce 同时预测所有输入多源大气观测的 computational complexity ，we performs iterative prediction in encoded low-dimensional latent space to 预测和降雨发生和演变真正相关的大气特征, 如图1所示.This framework fully leverages multi-source data 同时仅仅输出了关键的近地面雷达反射率因子.  而近地面反射率因子和地表降水率具有高度等价性。\cite{BERNE2004166}
% 1. To address the high complexity and susceptibility to noise and outliers in iterative prediction of multi-source data in the physical space, we construct a precipitation prediction framework that performs iterative prediction in encoded low-dimensional latent space. This framework fully leverages multi-source data while effectively reducing computational complexity by avoid.  

2. To capture the highly unbalanced distribution characteristics of precipitation events, we design a new loss function by combining pixel-wise MAE and cross-entropy loss, which increases penalties for missed detections while accurately constraining precipitation intensity values.  %没有超参数需要调整

3. The quantitative and qualitative comparisons conducted on two spatio-temporally heterogeneous datasets, with precipitation rate and radar reflectivity as respective prediction targets, comprehensively demonstrate the effectiveness and efficiency of the proposed forecasting model and loss function. Qualitative results further verify the model's superior capability in generating high-fidelity details.

\section{Related Work}

% Operational weather forecasting based on Numerical Weather Prediction (NWP) builds upon decades of research grounded in physical laws to simulate atmospheric behavior. Although the substantial growth in observational data, along with advances in science and computing, has improved forecast skill by approximately one day per decade \cite{Bauer2015}, the recent achievements of deep learning across scientific fields have prompted growing interest in its application to weather prediction \cite{gmd-17-2347-2024}\cite{atmos16010082}\cite{CampsValls2025}.

\subsection{Precipitation nowcasting (0-2 hours)}
In the domain of nowcasting, Prudden et al. \cite{prudden2020review} offer a comprehensive overview of radar-based nowcasting techniques and various machine learning approaches employed historically. And Davide et al. \cite{DeLuca2025} provided a concise review of four main rainfall nowcasting approaches: remote sensing-based extrapolation, numerical weather prediction, stochastic modeling, and deep learning techniques. In the field of precipitation nowcasting using deep learning, Shi et al. \cite{NIPS2015_07563a3f} introduced a recurrent neural network architecture utilizing convolutional LSTMs \cite{Graves2012}. Wang et al. developed PredRNN \cite{NIPS2017_e5f6ad6c}, which enhances spatiotemporal modeling with zigzag memory flow and decoupled memory cells. Gao et al. introduced Earthformer \cite{NEURIPS2022_a2affd71}, a cuboid-based Transformer for Earth system prediction, and later proposed PreDiff \cite{NEURIPS2023_f82ba6a6}, a two-stage diffusion model with physical constraints for probabilistic forecasting.

In addition. Agrawal et al. \cite{agrawal2019machine} framed the forecasting task as an image-to-image translation problem using a U-Net model. Trebing et al. \cite{trebing2021smaat} incorporated attention mechanisms, demonstrating a reduction in model parameters without sacrificing performance. Ravuri et al. \cite{Ravuri2021} developed a deep generative model for probabilistic nowcasting, achieving realistic predictions up to 90 minutes and improving medium-to-heavy rainfall forecasts. Zhang et al. \cite{Zhang2023} proposed NowcastNet, which integrates physical principles through a deterministic evolution network and a stochastic generator to predict extreme precipitation at kilometer scales. However, these methods primarily focus on the nowcasting of severe convection and thunderstorms, and their skillful forecast time is mostly limited to 0--3 hours.

\subsection{Precipitation forecasting (exceeding 2 hours)}
In short term forecasting, Sønderby et al. proposed Metnet \cite{snderby2020metnet}, a neural weather model based on CNNs and RNNs, which provides probabilistic precipitation forecasts up to 8 hours ahead at 1 km spatial and 2 min temporal resolution. The model uses axial self-attention to efficiently aggregate global meteorological context from input areas spanning one million square kilometers, achieving both high accuracy and low latency. Building on this, Espeholt et al. developed Metnet2 \cite{Espeholt2022}, which extends the forecast range to 12 hours with the same spatiotemporal resolution. It incorporates enhanced architectural components to effectively process an even larger input context of 2048 km × 2048 km, improving the representation of large-scale weather patterns. Further advancing the series, Andrychowicz et al. introduced Metnet3 \cite{andrychowicz2023}, which significantly expands capability to a 24-hour lead time while adding predictions for multiple variables--including wind, temperature, and dew point. A key innovation is its densification technique, which generates spatially continuous forecasts from sparse observational inputs, markedly improving utility for operational forecasting. Despite these advances, such studies consume enormous computational resources, making it extremely challenging to tune hyperparameters or optimize network architectures.

In longer-range forecasting, deep learning has also been utilized for seasonal prediction of weather events, including extremes \cite{Ham2019}\cite{Yan2020}. For example, Chen et al. introduced FuXi Subseasonal-to-Seasonal (FuXi-S2S) \cite{Zhong2024}\cite{Chen2024}, a machine learning model inspired by VAE for global daily mean forecasts up to 42 days. Trained on 72 years of ECMWF ERA5 data, FuXi-S2S outperformed existing models, enhancing predictions for precipitation and the Madden–Julian Oscillation (MJO).

\section{Methods}
\subsection{Overall Framework}

\begin{figure*}[t]
\centering
    \includegraphics[width=1\textwidth,trim=0 60 30 0,clip]{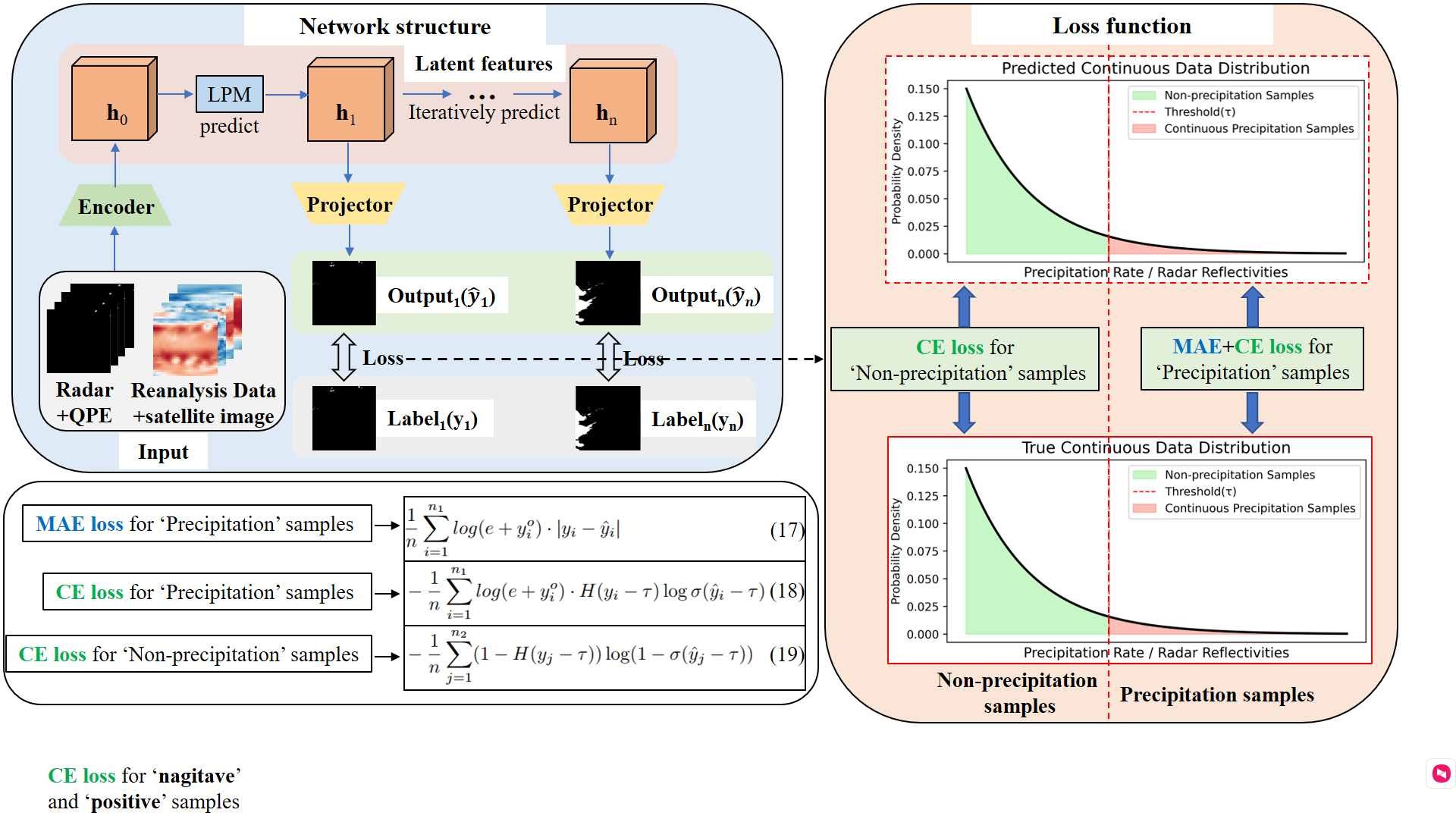}
    \caption{Network structure and loss function of our proposed forecasting model. The ‘labels’ include the MRMS hourly cumulative QPE from ‘USA’ dataset or near-surface radar reflectivity from ‘Hubei’ dataset. And the ‘MAE’ and ‘CE’ losses for precipitation samples employ a weighting scheme, as presented in Eq. \ref{a9990} and Eq. \ref{a9991}. ‘LPM’ represents the latent feature iterative prediction model. It is worth noting that the encoder simultaneously encodes the latent features from two time steps.}
    \label{fig:network}
\end{figure*}
To enable iterative prediction in a low-dimensional latent space, we design an encode--iterative predict--project framework. As illustrated in Fig. \ref{fig:network}, the model encodes all input atmospheric variables from two time steps, performs iterative prediction in the latent space, and finally projects the latent feature to obtain the predicted results. 
This framework efficiently integrates and utilizes multi‑source data by iteratively predicting only the latent features strongly associated with precipitation, thereby avoiding the need to forecast all input atmospheric variables. The projected outputs are precipitation rates \cite{MRMS} or the near‑surface radar reflectivity factor, the latter of which exhibits strong equivalence to precipitation rates \cite{BERNE2004166}.
\subsection{Forecasting Model}
\subsubsection{Encoder}
\begin{figure*}[t]
\centering
    \includegraphics[width=0.8\textwidth,trim=0 240 250 0,clip]{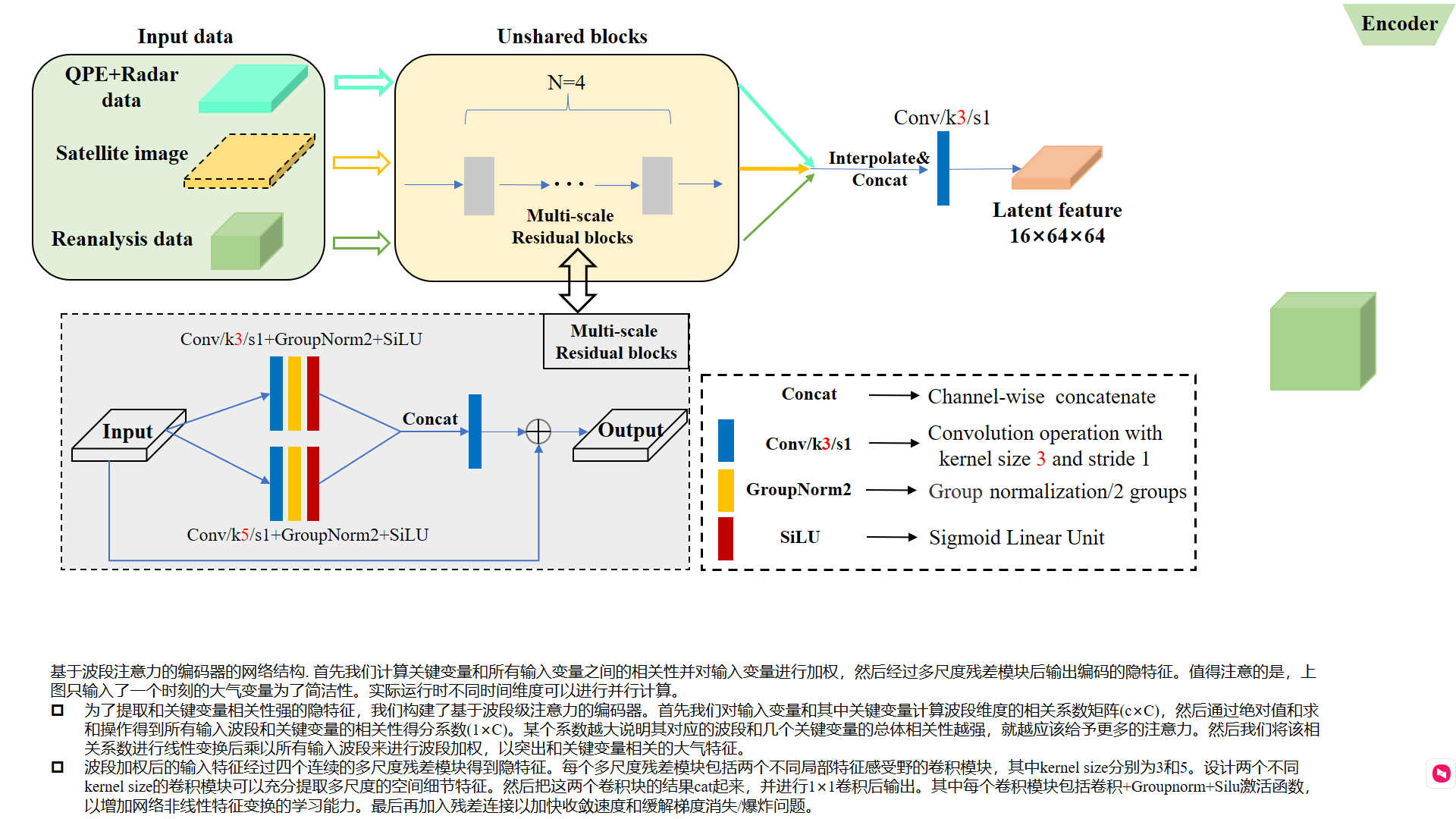}
    \caption{Network structure of the designed encoder. Note that the input for the ‘USA’ dataset does not include satellite imagery and the input for the ‘Hubei’ dataset does not include MRMS QPE.}
    \label{fig:encoder}
\end{figure*}

As shown in Fig. \ref{fig:encoder}, we design an encoder to extract latent features from multi-source atmospheric observations including MRMS QPE (if have) and weather radar data ($x\_{qpe+radar}$), ERA5 reanalysis data ($x\_{reanalysis}$) \cite{era5_bibtex}, and geostationary satellite imagery ($x\_{satellite}$, when available):
\begin{equation}
\label{a02f}
\begin{aligned}
    h=\varepsilon(x_{qpe+radar}, x_{reanalysis}, x_{satellite})
\end{aligned}
\end{equation}
\noindent where $\varepsilon()$ represents the encoder. In this research, the dimension of encoded latent feature is set to 16×64×64 (channel×height×width). The extracted latent features are guided by loss functions strongly associated with precipitation initiation and evolution.

As shown in Fig. \ref{fig:encoder}, the encoder first processes each data modality through four dedicated multi-scale residual blocks for modality-specific feature extraction. Notably, the feature extractors are not shared across different atmospheric observation variables. Each multi-scale residual block contains two parallel convolutional modules with different kernel sizes, which enables comprehensive extraction of multi-scale features from the input atmospheric variables. Subsequently, these features are fused to adaptively weigh the importance of different scales:
\begin{equation}
\begin{split}
\text{MultiScaleBlock}(x) = &x + \text{Conv}_{1\times1}\left(\text{Cat}\left[ 
    \begin{aligned}
        &\text{ConvBlock}_{3\times3}(x), \\
        &\text{ConvBlock}_{5\times5}(x)
    \end{aligned}
    \right] \right)
\end{split}
\end{equation}
\noindent where $\text{ConvBlock}_{3\times3}$ includes a convolutional layer with a kernel size of 3×3, a Group Normalization divided into two groups, and a SiLU activation function. ‘Cat’ indicates the concatenation operation along the channel dimension.

The extracted atmospheric features from the multi-source observations finally undergo spatial interpolation for dimensional alignment, channel-wise concatenation of all features, and a final 3×3 convolutional layer to produce the latent feature \( h \).
% As shown in Fig. \ref{fig:encoder}, each multi-scale residual block consists of two convolutional modules with the kernel sizes of 3 and 5, respectively. Designing convolutional modules with two different kernel sizes allows for the full extraction of multi-scale spatial features.
% Note that $m_y > m_h > m$. In this study, the latent feature dimension $m_h$ was experimentally set to 32. 
% For the global dataset, the spatial dimensions of the input variables and the latent feature are consistent, whereas for the regional dataset, the spatial dimensions of the input variables are larger to account for boundary conditions, while the spatial dimensions of the latent feature are same as prognostic variables'.

\subsubsection{Latent Dynamics Predictor (LPM)}
\begin{figure*}[t]
% \vskip 0.01in
\centering
    \includegraphics[width=1\textwidth,trim=20 300 380 0,clip]{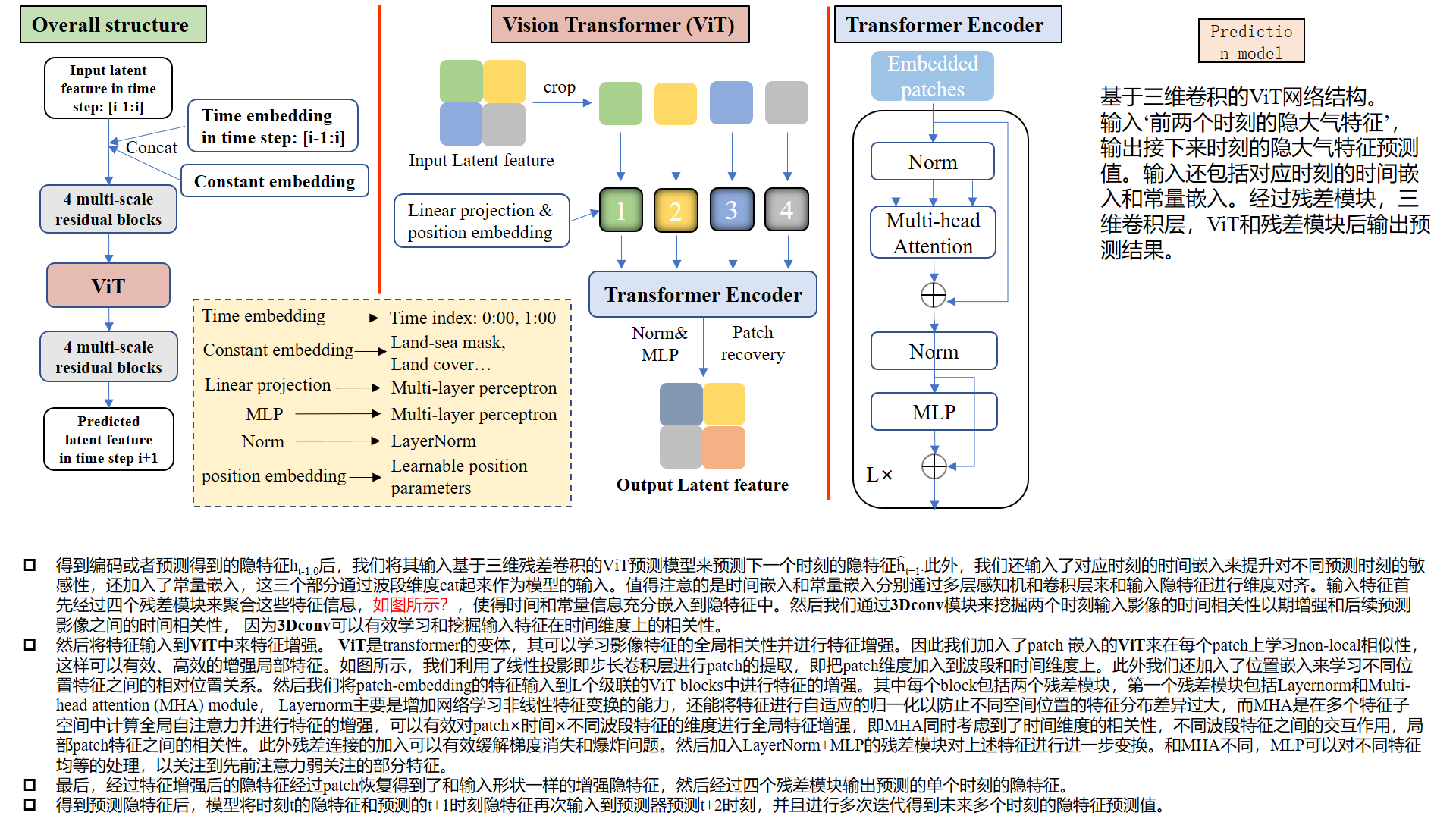}
    \caption{The network structure of ViT-based latent feature iterative prediction model ($LPM$). 
    It takes the latent atmospheric features of the previous two time steps as input and outputs the predicted latent feature for the next time step. The input also includes the corresponding time embeddings and constant embeddings. After passing through the multi-scale residual blocks, 3D convolutional layer, ViT, and the multi-scale residual blocks again, the predicted results are output. Note that the ‘multi-scale residual blocks’ are identical to those in the encoder, as depicted in Fig. \ref{fig:encoder}.
}
    \label{fig:LPM_network}
% \vskip -0.3in
% \vskip -0.2in
\end{figure*}
\begin{figure*}[t]
% \vskip 0.01in
\centering
    \includegraphics[width=0.8\textwidth,trim=50 180 50 40,clip]{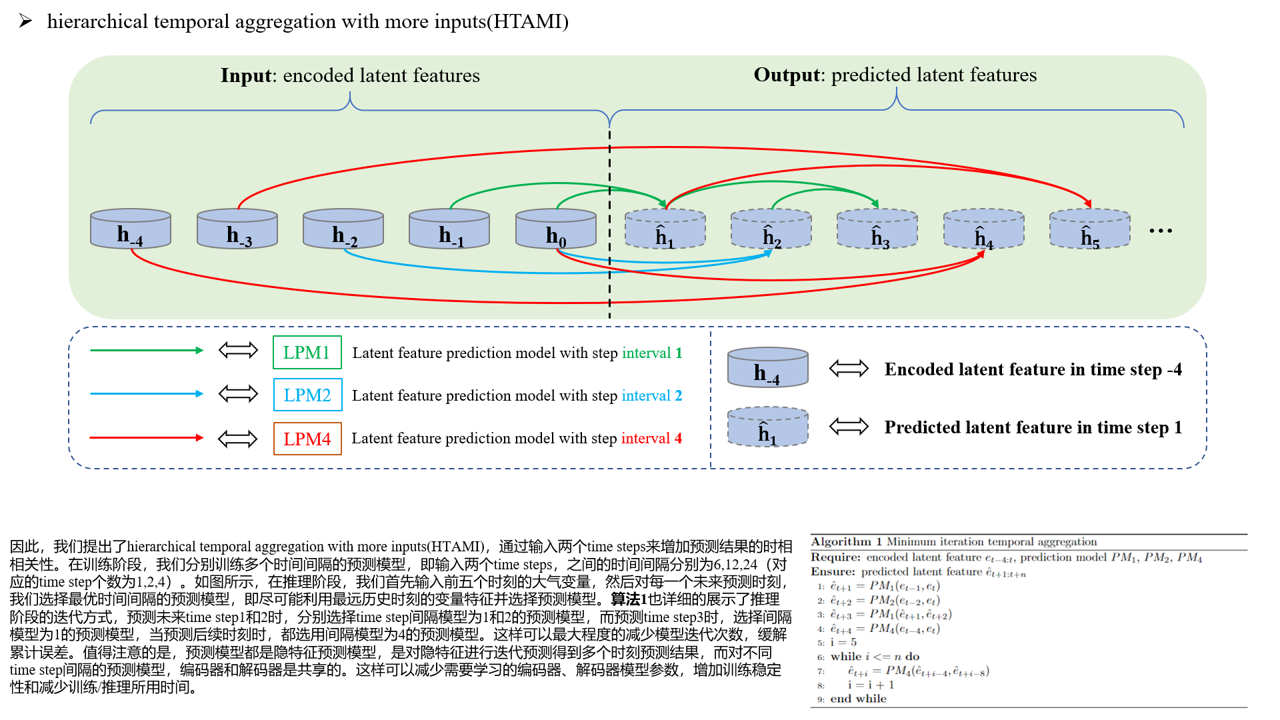}
    \caption{A schematic diagram of the inference process for the improved HTA algorithm, which takes two time steps as input for iterative prediction. Arrows of different colors represent predictions with different time step intervals.
}
    \label{fig:HTAMI}
% \vskip -0.3in
% \vskip -0.2in
\end{figure*}
% \vspace{0.3pt}
After obtaining the encoded latent features $h_{-1:0}$ (where the subscript denotes time steps $-1$ and $0$), we feed them into a Vision Transformer (ViT)-based Latent Feature Iterative Prediction Module (LPM) to predict the latent feature at the next time step--$\hat{h}_1$. Additionally, we include the time embedding in time steps $-1:0$ ($t_{-1:0}$) to enhance the sensitivity to different input times and add the constant embedding $c$. These three components are concatenated along the channel dimension to serve as input of $LPM$:
\begin{equation}
\label{a02d}
\begin{aligned}
    \hat{h}_1=LPM(h_{-1:0}, t_{-1:0}, c) 
\end{aligned}
\end{equation}
\begin{equation}
\label{a02e}
\begin{aligned}
    \hat{h}_{i+1}=LPM(\hat{h}_{i-1:i}, t_{i-1:i}, c), i=1...L-1
\end{aligned}
\end{equation}
\noindent where $h, \hat{h}, t_{i-1:i}$ means the encoded and predicted latent features and time embeddings in time steps $i-1$ and $i$. For longer-range forecasts, the $LPM$ is unfolded by iteratively feeding its predictions back into the model as input and $L$ means the number of total prediction time steps which is set to 24 in this study. 

To mitigate the accumulation of errors caused by multiple iterations, we incorporate the HTA algorithm \cite{bi2023accurate}, which involves three $LPM$ units that predict latent features at time step intervals of 1, 2, and 4, respectively. 
It is worth noting that we have improved the original HTA algorithm by introducing two time steps with different intervals during the training phase to provide more historical atmospheric information. During the inference phase, we input five time steps to iteratively predict the latent features for future time steps. Specifically, as shown in Fig. \ref{fig:HTAMI}, the latent features for the first and second future time steps are predicted by LPMs with intervals of 1 and 2 time steps, respectively, while the latent feature for the third lead time step is predicted by the LPM with an interval of 1 time step. Starting from the fourth lead time step, the latent features for all subsequent time steps are predicted by the LPM with an interval of 4 time step. This strategy maximizes the use of the three LPMs with different intervals to mitigate error accumulation.

% During the inference phase, as shown in Fig. \ref{fig:HTAMI}, we input five time steps to iteratively predict the latent features for future time steps. Specifically, the latent features for the first and second future time steps are predicted by LPMs with intervals of 1 and 2, respectively, while the latent feature for the third lead time step is predicted by the LPM with an interval of 1. Starting from the fourth lead time step, the latent features for all subsequent time steps are predicted by the LPM with an interval of 4. This strategy maximizes the use of the three LPMs with different intervals to mitigate error accumulation.

As shown in Fig. \ref{fig:LPM_network}, input latent features first pass through four multi-scale residual blocks (as in Fig. \ref{fig:encoder}) to aggregate different feature information including $h, t$ and $c$.
% as shown in the left part of Fig. \ref{fig:LPM_network}, 
% allowing time and constant information to be fully embedded into the latent features. 
% We then employ a 3-dimensional convolution (3Dconv) module to mine the temporal correlation between two time steps’ input latent features, aiming to enhance the temporal correlation with the predicted latent feature.
% $h \in \mathbb{R}^{m_h\times pq}$ 
The latent features are then fed into ViT for feature enhancement and prediction. ViT \cite{dosovitskiy2020image} is a variant of transformer \cite{NIPS2017_3f5ee243} that can learn the global correlation of image features and enhance them. And it introduce the patch embedding to efficiently learn the patch-wise non-local attention to enhance the patch features. Finally, enhanced latent features are passing through four multi-scale residual blocks to output the predicted latent feature in next time step.

As shown in Fig. \ref{fig:LPM_network}, the network architecture of the ViT mainly consists of the following components. First, a linear projection layer is used to divide the input latent features into patches, where the patch size $P\times P$ in this study is set to 4×4:
\begin{equation}
    \mathbf{Z}_0 = \left[\mathbf{h}^1_p; \mathbf{h}^2_p; \cdots; \mathbf{h}^N_p \right] + \mathbf{E}_{\text{pos}}
\end{equation}
where $\mathbf{h}_p^i \in \mathbb{R}^{P^2 \times C}$ denotes the $i$-th patch from linear projection, and $\mathbf{E}_{\text{pos}}$ is the positional encoding. $C$ denotes the number of latent feature channels, which is set to 16 in this study. The resulting patches, combined with positional encoding, are then fed into the transformer encoder. This encoder consists of $L$ cascaded Transformer\_Blocks designed to learn and leverage global correlations for predicting the latent features of the next time step:
% Transformer Encoder with L Blocks
\begin{equation}
    \mathbf{Z}_\ell = \text{Transformer\_Block}(\mathbf{Z}_{\ell-1}), \quad \ell = 1, 2, \dots, L
\end{equation}
where each Transformer\_Block contains the following network layers:
% Single Transformer Block
\begin{equation}
\begin{aligned}
    \mathbf{Z}'_\ell &= \text{MHA}\left(\text{Norm}(\mathbf{Z}_{\ell-1})\right) + \mathbf{Z}_{\ell-1} \\
    \mathbf{Z}''_\ell &= \text{Norm}(\mathbf{Z}'_\ell) \\
    \mathbf{Z}_\ell &= \text{MLP}\left(\mathbf{Z}''_\ell\right) + \mathbf{Z}''_\ell
\end{aligned}
\end{equation}
where ‘MHA’ denotes multi-head self-attention, ‘Norm’ refers to the LayerNorm layer, and ‘MLP’ is a two-layer feed-forward network with GELU activation. Finally, the output of the ViT model is obtained through the Patch Recovery and related operations:
% Output Processing
\begin{equation}
\begin{aligned}
    \mathbf{z} &= \text{MLP}\left(\text{Norm}(\mathbf{Z}_L)\right) \\
    \hat{h} &= \text{Patch\_Recovery}(\mathbf{z})
\end{aligned}
\end{equation}

\subsubsection{Projector}
The latent features generated through iterative forecasting across multiple lead times are transformed into predicted precipitation rates or radar reflectivity values via parallel projection. The projector network architecture comprises a ‘multi-scale residual block’ (same as that in encoder) and a patch-based multi-head attention layer (same as that in ViT and with patch size=4×4 and 2 attention heads) \cite{NIPS2017_3f5ee243}, as illustrated in Figs. \ref{fig:encoder} and \ref{fig:LPM_network}. The processed features are then upsampled to the target resolution, and finally passed through a convolutional layer with kernel size 3 to produce the output predictions:
\begin{equation}
\label{a_dec}
\begin{aligned}
    \hat{y}_i=\mathfrak{P}(\hat{h}_{i}), i=1...L
\end{aligned}
\end{equation}
\noindent where $\mathfrak{P}$ represents the projector. Note that the projector implements parallel computation over multiple lead times.

\subsubsection{Reconstructor}  % thereby falling into a local minimum of the predictor
Due to the limited number of iterations during the training phase, to prevent the encoder from extracting features that are only easily predictable within a few time steps, we design a reconstructor. Specifically, the encoded latent features should be as informative as possible, enabling the LPM to learn a crucial atmospheric state representation's transitions, thereby ensuring stable long-term prediction performance. The reconstructor network architecture for each modal feature ($x\_{qpe+radar}, x\_{reanalysis}$ and $x\_{satellite}$) consists of two convolutional layers with a LeakyReLU activation function in between:
\begin{equation}
\label{a_recons}
\begin{aligned}
    \hat{x}_i=\mathfrak{R}(\hat{h}_{i}), i=-1, 0...L
\end{aligned}
\end{equation}
\noindent where $\mathfrak{R}$ represents the reconstructor. Note that the reconstructor also implements parallel computation over multiple lead times.

\subsection{Loss Functions}
% \begin{figure*}[ht]
% \centering
%     \includegraphics[width=0.8\textwidth,trim=0 0 0 0,clip]{picture/general/rain_pixels_vs_loss_nexrad_shift.png}
%     \caption{The standard cross-entropy loss curve ($\mathcal{L}_{CE}$) and the newly designed loss function curve ($\mathcal{L}_{ELT-CE}$) for 0-1 normalized radar reflectivity values, along with the data distributions above the threshold $\tau$. $\tau$ represents the 0-1 normalized threshold derived from the original radar reflectivity's precipitation threshold (usually is 20 dBZ).
%     The newly designed loss function $\mathcal{L}_{ELT-CE}$ could accurately constrains precipitation samples of varying intensities by adaptively adjusting its zero point (via the shift term $s$) and increase the penalty for the missed precipitation samples.
% }
%     \label{fig:rain_pixels_vs_loss}
% \end{figure*}
\subsubsection{The proposed ‘WMCE’ loss function}
First, it is demonstrated that the pixel-wise MAE/MSE loss tends to predict precipitation events as non-precipitation events. This is because the pixel-wise MAE/MSE loss function primarily focuses on high-probability samples (non-precipitation events, negative samples) while neglecting extremely low-probability samples (precipitation events, positive samples). Second, results demonstrate that by imposing a heavier penalty on extremely scarce precipitation events, the proposed WMCE loss significantly improves the accuracy of both precipitation event discrimination and intensity prediction.
% (Notably, for numerical stability, the radar echo intensity values in this paper are normalized to the range of [0, 1]).
% and the precision of rainfall intensity regression.

(1) The commonly used pixel-wise losses $MSE(Y, \hat{Y})$ and $MAE(Y, \hat{Y})$ between the predicted precipitation intensity value $\hat{Y}$ and its truth $Y$ can be formulated as:
% \begin{equation}
% \label{a05}
% \begin{aligned}
%     argmin \; MSE(Y, \hat{Y}) = argmin \; -\frac{1}{n}\log (q_\theta(y_1)q_\theta(y_1)...q_\theta(y_n))
% \end{aligned}
% \end{equation}
% \noindent $q_\theta$ outputs the predicted precipitation intensity distribution, typically assumed to be a Gaussian distribution \( N(\hat{y}_i, \hat{\sigma}_i) \). Note that the log transformation is usually applied to reduce computational complexity and optimization difficult:
% \begin{align}
%     MSE(Y, \hat{Y}) &= -\frac{1}{n}\sum_{k=1}^n\log q_\theta(y_k) \label{a060} \\ & = -\frac{1}{n}(\sum_{i=1}^{n_1}\log q_\theta(y_i) - \sum_{j=1}^{n_2}\log q_\theta(y_j)) \label{a061}
%     \\ & = -\frac{1}{n}(\sum_{i=1}^{n_1}(\log c_1-\frac{(y_i-\hat{y}_i)^2}{2\hat{\sigma}^2_i}) - \sum_{j=1}^{n_2}(\log c_2-\frac{(y_j-\hat{y}_j)^2}{2\hat{\sigma}^2_j})) \label{a062}
% \end{align}
\begin{equation}
\label{a001}
\begin{aligned}
    MSE(Y, \hat{Y}) = \frac{1}{n}(\sum_{i=1}^{n_1}(y_i-\hat{y}_i)^2 + \sum_{j=1}^{n_2}(y_j-\hat{y}_j)^2)
\end{aligned}
\end{equation}
\begin{equation}
\label{a0011}
\begin{aligned}
    MAE(Y, \hat{Y}) = \frac{1}{n}(\sum_{i=1}^{n_1}|y_i-\hat{y}_i| + \sum_{j=1}^{n_2}|y_j-\hat{y}_j|)
\end{aligned}
\end{equation}
\noindent where $n_1$ counts grid points with precipitation event happen and $n_2$ counts grid points without precipitation event ($n_1 + n_2 = n$). Note that $n_1 \ll n_2$, which means the non-precipitation events accounts for a significantly large proportion, as shown in Table \ref{tab:tab0}. 
From the above equation, it can be inferred that since \( n_1 \ll n_2 \), the optimization process primarily focuses on minimizing the objective function \( \sum_{j=1}^{n_2}(y_j - \hat{y}_j)^2 \) or $\sum_{j=1}^{n_2}|y_j-\hat{y}_j|$, thereby neglecting the minimization of another term. Consequently, this leads to substantial under-detection of precipitation events and inaccurate quantification of predicted precipitation intensity.
% 为了稳定网络输出，我们需要对雷达回波强度数据进行归一化，由于雷达强度0值太多，导致方差太小非常接近0，如果使用标准归一化即减去均值除以方差，可能导致数值范围太大。因此本研究将雷达回波强度范围归一化到了0-1。

% \begin{figure*}[t]
% \vskip 0.01in
% \centering
% \includegraphics[width=1\textwidth,trim=50 50 20 100,clip]{picture/general/two_distributions.png}
%     \caption{Two distributions}
%     \label{fig:two_distributions}
% \vskip -0.2in
% \end{figure*}

(2) To mitigate the aforementioned limitations of MSE/MAE losses, we adopt a cross-entropy (CE) loss to separate precipitation samples (labeled as 1) from non-precipitation samples (labeled as 0). The cross-entropy loss is highly sensitive to misclassifications--even for rare events--due to the sharp increase in the log function’s output as its input approaches zero.

The precipitation values can then be precisely constrained using a MAE/MSE loss following correct class separation. In contrast, non-precipitation samples only require categorical discrimination and do not require this precise regression computation, as shown in right part of Fig. \ref{fig:network}.
This design reduces the complexity of the learning task and mitigates the risk of overemphasizing non-precipitation samples when a pixel-wise MAE/MSE loss is applied across all samples. Furthermore, omitting the MAE/MSE loss for non-precipitation samples does not compromise the accuracy of our latent-space iterative forecasting framework, since precipitation predictions at each time step depend solely on the corresponding latent features and are independent of earlier precipitation predictions. It should be noted, that this loss function is not suitable for models that perform iterative prediction directly in the raw physical variable space, as it would significantly impair forecasting accuracy.

To fully align with the characteristics of the log function in CE loss function and to improve stability, the target variables--hourly cumulative precipitation rates or radar reflectivity values--are normalized to the range [0, 1] prior to training.
%我们利用了交叉熵损失函数来分离有雨(标签为1)和无雨(标签为0)强度值。%交叉熵损失函数对于误分现象非常敏感，因此能够提高对发生概率小的事件的预测能力。这是由于交叉熵损失函数中log函数的非线性变化的特性。
% 分离开后，有雨事件再利用回归loss即可达到较高的降雨强度预测精度。而无雨事件的强度值只需要和有雨事件的区分开就行，并不值得回归预测。

% cross-entropy ($CE$) loss function的公式为：
Firstly, the cross-entropy (CE) loss function for binary classification could be formulated as:
\begin{equation}
\label{a0701}
\begin{aligned}
CE(p, q_\theta) = -\mathbb{E}(p\log q_\theta + (1-p)\log (1-q_\theta))
\end{aligned}
\end{equation}
\noindent Where the label $p = 1$ indicates a precipitation event. When precipitation rates or radar reflectivity values exceed a predefined threshold $\tau$, they are classified as precipitation samples/events ($p = 1$); otherwise, they are treated as non-precipitation events ($p = 0$). The term $q_{\theta}$ represents the probability of precipitation occurrence. Accordingly, the cross-entropy loss can be reformulated as follows:
\begin{align}
\mathcal{L}_{CE}(Y, \hat{Y}) &= CE(H(Y-\tau), \sigma(\hat{Y}-\tau)) \label{a990} \\
\begin{split}
&= -\frac{1}{n}\Biggl(\sum_{i=1}^{n_1} p_i\log \sigma(\hat{y}_i-\tau) \\
&\qquad +\sum_{j=1}^{n_2} (1-p_j)\log (1-\sigma(\hat{y}_j-\tau))\Biggr) \\
\end{split} \\
\begin{split}
&= -\frac{1}{n}\Biggl(\sum_{i=1}^{n_1} H(y_i-\tau)\log \sigma(\hat{y}_i-\tau) \\
&\qquad +\sum_{j=1}^{n_2} (1-H(y_j-\tau))\log (1-\sigma(\hat{y}_j-\tau))\Biggr) \label{a991}
\end{split}
\end{align}
\noindent where $p$ in $CE$ loss function is computed as $H(Y-\tau)$ which is the \textbf{H}eaviside Step Function and it return label $p=1$ if precipitation intensity label $Y\geq \tau$ else return label $p=0$. $q_\theta$ is computed as $\sigma(\hat{Y}-\tau)$ where $\hat{Y}$ is the predicted precipitation intensity and $\sigma()$ means the sigmoid function to output the ‘probability’ of precipitation event happen. 
% And $\sigma(\hat{Y} - \tau)$ can drive the intensity values of precipitation events greater, thereby preventing them from being predicted as the no-rain events with the intensity values less than $\tau$. 
% 让我们通过一个简单的例子来说明为什么ce loss对小概率事件的错误预测会很敏感。假设1个有雨像素都被错误的预测为了无雨像素，即FN，并且上式子中的$q_{\theta,i}$约为0.02，因为被预测为无雨所以其sigmoid输出值接近0，那么log(0.02)约等于-3.912。而如果利用mse或者l1 loss,那么loss值最多为1（当将数值范围归一化到0-1时）。可以看出ce loss对于错分类的像素值更加敏感。
% 然而，如图所示，如果我们在原始空间中使用如式8的损失函数，交叉熵损失函数值在大约25后就非常接近0了，即像素反射率值大于25的降雨事件很难被检测到。而在0-1归一化的像素空间中，原始交叉熵loss又在0-1内一直大于0，会驱动着有雨事件的强度值一直增大，导致了预测的雷达反射率的高估。此外，当有雨事件误分为无雨时，即预测强度值小于阈值\tau 时，我们希望loss变得更大，对其惩罚更强。因此我们设计了新的基于交叉熵的损失函数：
% ***
% 然而，交叉熵损失函数仍然遭受着只关注高概率无降水样本的问题，导致对有降水样本的漏检。通过增加降水样本的比例可以降低漏检率但会增加误检率，在实际应用时会导致预测的降水事件比例的增加。此外，如图中蓝线所示（即原始交叉熵损失函数）该loss在0-1范围内缓慢减小且一直大于0，可能导致预测降水强度值的高估。因此我们设计了如下损失函数来增加对漏检降水样本的惩罚：
% In some studies, by increasing the proportion of precipitation samples can reduce the miss rate, but it simultaneously increases the false alarm rate - a trade-off that would artificially inflate the predicted precipitation frequency in practical applications.  
Actually, the 'probability' in Equation \ref{a991} represents the distance between the precipitation intensity value and the threshold $\tau$. Although this loss function remains positive within the normalized [0, 1] range--which may introduce an upward bias in the predicted precipitation intensity values--it can still effectively constrain precipitation intensity by combining with MAE/MSE loss. Furthermore, the conventional CE loss ($\mathcal{L}_{CE}$) still suffers from overemphasizing high-probability non-precipitation events, resulting in missed detections of precipitation events.
To address these limitations, we introduced \textbf{w}eighting to both the original pixel-wise \textbf{M}AE loss and the \textbf{CE} loss (‘WMCE’), such that higher precipitation intensities are assigned greater loss weights (The MAE loss function was chosen for its ability to mitigate blurry effects):
\begin{align}
\mathcal{L}_{\mathrm{WMCE}} = & \frac{1}{n}\sum_{i=1}^{n_1}log(e+y_{i}^o) \cdot |y_i-\hat{y}_i| \label{a9990} \\& -\frac{1}{n}\sum_{i=1}^{n_1} log(e+y_{i}^o) \cdot H(y_i-\tau)\log \sigma(\hat{y}_i-\tau) \label{a9991} \\&-\frac{1}{n} \sum_{j=1}^{n_2} (1-H(y_j-\tau))\log (1-\sigma(\hat{y}_j-\tau))
\label{a9992}
\end{align}
\noindent 
% where $y_{i}^o$ 代表了原始未归一化的Instantaneous precipitation rates或者雷达反射率因子，$y_{i}$是归一化后的真值。$e$指自然指数，使得降水样本的最小权重大于1。Eq \ref{a9990} 代表了只包含降水样本的加权MAE loss，Eq \ref{a9991} 代表了只包含降水样本的加权CE loss，Eq \ref{a9992} 代表了只包含无降水样本的无加权的CE loss。再次强调, precipitation rates or reflectivity values below the threshold $\tau$ were excluded from the MAE loss calculation. This exclusion was implemented because it effectively mitigated overattention to non-precipitation events when calculate pixel-wise loss over all samples and 降低了学习任务的复杂度。并且不会影响所构建的隐空间迭代预测模型的迭代预测精度，因为各个时刻的降水预测结果仅和对应时刻的隐特征相关，和之前时刻降水预测结果无关。
where \( y_{i}^o \) represents the original unnormalized truth hourly cumulative precipitation rate or radar reflectivity factor, and \( y_{i} \) is the corresponding normalized ground truth. The Euler's number \( e \) ensures that the minimum weight for precipitation samples is greater than 1. Equation \ref{a9990} defines the weighted MAE loss applied only to precipitation samples, Equation \ref{a9991} corresponds to the weighted CE loss for precipitation samples, and Equation \ref{a9992} denotes the unweighted CE loss for non-precipitation samples. 

\subsubsection{Latent Feature and Reconstruction Loss}
(1) We also constrain the predicted latent features to closely match the true latent features by applying an L1 loss function between the latent features predicted by the LPM and the true latent features encoded from the input variables at the corresponding time steps:
\begin{equation}
\label{a02a}
\begin{aligned}
    \mathcal{L}_{latent} = L1\_loss(h_i, \hat{h}_i), \quad i=1,2...L_t
\end{aligned}
\end{equation}
\noindent where $\hat{h}_i$ is the predicted latent feature at time step $i$ and $h_i$ is its truth. $L_t$ is the number of iterative prediction steps in training stage, which is set to 2.

(2) In addition, we incorporate a reconstruction loss between the true physical variables $x_i$ (including $x\_{qpe+radar}$, $x\_{reanalysis}$ and $x\_{satellite}$ (if have)) and reconstructed physical variables $\hat{x}_i$ that directly reconstructed from the encoded latent feature:
% By combining with $\mathcal{L}_{\mathrm{WMCE}}$, this loss enables the encoder to distill latent atmospheric features that are strongly associated with precipitation from a vast set of input variables: %过度追求在少数几个步数内让预测器容易预测。
\begin{equation}
\label{a02c}
\begin{aligned}
    \mathcal{L}_{recon} = L1\_loss(x_i, \hat{x}_i), \quad i=-1,0...L_t
\end{aligned}
\end{equation}

\noindent Therefore, the overall loss function is the combination of above loss functions:
\begin{equation}
\label{MSE_all}
\begin{aligned}
\mathcal{L}_{\mathrm{overall}}= \mathcal{L}_{\mathrm{WMCE}} + \mathcal{L}_{\mathrm{latent}} + \mathcal{L}_{\mathrm{recon}}
\end{aligned}
\end{equation}
\noindent Note that all loss functions are assigned a weight of 1, except for the reconstruction loss of ERA5 variables, which has a weight of 0.1. 
% To place greater emphasis on heavy precipitation, the MAE loss is further weighted by the logarithm of the original unnormalized radar reflectivity values, denoted as $\log(y_{i} \in y_\text{original})$. Importantly, reflectivity values below the threshold $\tau$ were excluded from the MAE loss calculation. This exclusion was implemented because it effectively mitigated overattention to non-precipitation events when regressing over the full reflectivity range. Notably, this exclusion did not compromise the accuracy of iterative predictions; on the contrary, it aligned well with our latent space iterative prediction framework.

\section{Results}\label{experiment}
% This section introduce the datasets, experimental details, compared methods, comparative results and ablation studies.
\subsection{Dataset}\label{sec_Data}
% maxlat, minlat, minlon, maxlon = 57, 18, 90, 126
\begin{table*}[t]
% \vskip 0.01in
\caption{The details of two datasets. ERA5 variables include 6 upper-air variables in 37 levels and 21 surface variables. 6 upper-air variables include ‘geopotential’, ‘temperature’, ‘u component of wind’, ‘v component of wind’, ‘specific humidity’ and ‘vertical velocity’. Please refer to the supplementary materials for more details. The time resolution of these two datasets are 1 hour and 15 minutes, respectively.}\label{tab_dataset}%
\setlength{\tabcolsep}{2pt}
\begin{center}
\begin{small}
% \begin{sc}
\begin{tabular}{cccccc}
\toprule
Dataset & Region & Input Variables & Spatial & Predicted variable & Time \\
name & & (2 time step) & resolution & (24 time steps) & range\\
\midrule
USA & 100.9°W-83.1°W ‌ & ERA5 variables(243) & 0.25°(77×73) & Cumulative  & 2015.01.01- \\
dataset & 27.6°N-46.4°N & 3D Nexrad reflectivity(24) & 6km(256×256) & MRMS QPE & 2018.01.01 \\
 & & Cumulative MRMS QPE (1) & 6km(256×256) & (1×256×256) & \\
\midrule
Hubei & 111.9°E-116.8°E & ERA5 variables(243) & 0.25°(37×37) & Near ground radar & 2022.6.1-10.1   \\ 
dataset & 28.3°N-32.7°N & FY4B satellite L1 data(15) & 4km(256×256) & reflectivity & 2023.5.1-10.1 \\
 & & 3D radar reflectivity(9) & 2km(250×250) & (1×250×250) &  \\
 % U10m, V10m, T2m, mslp, sp, TCWV
 % T2m, mslp, sst, U100m, V100m
 % [100.94, 83.06, 27.56, 46.44]
\bottomrule
\end{tabular}
% \end{sc}
\end{small}
\end{center}
% \vskip -0.2in
\end{table*}
% \textcolor{red}{This is a fair concern, and we acknowledge the limitations of using only five variables and a 5.625° resolution. However, these choices were driven by the computational constraints faced by our research team. Training a neural ODE model at 1.5° or 0.25° resolution requires substantial resources—potentially hundreds or thousands of GPUs—making it feasible only for major tech companies like Google, Meta, or NVIDIA. For instance, Google's GraphCast achieves 0.25° resolution without neural ODEs, and even their neuralGCM model is limited to 1.4° resolution due to the computational demands.}
The validity of the proposed prediction model was verified through experiments conducted on two datasets, with detailed specifications provided in Table \ref{tab_dataset}. 
Note that the Cumulative Quantitative Precipitation Estimation (QPE) products in Table \ref{tab_dataset} are derived from radar reflectivity through empirical conversion formulas \cite{MRMS}. Furthermore, most studies indicate that a radar reflectivity exceeding 20 dBZ corresponds to the presence of precipitation droplets, while a threshold of 0.2 mm/h is also widely considered indicative of measurable precipitation \cite{Espeholt2022}. Therefore, precipitation samples greater than 0.2 mm/h and radar reflectivity samples exceeding 20 dBZ are highly equivalent, both representing near-surface precipitation intensity (the threshold $\tau$ in the designed loss function is also set to 0.2 mm/h and 20 dBZ). Table \ref{tab:tab0} also shows that both types of samples constitute only a very small fraction of the total dataset. Therefore, the prediction tasks for these two targets (QPE and radar reflectivity) are highly similar, which justifies the feasibility of forecasting both with a single model.

The prediction includes 24 lead time steps. Since the temporal resolutions of the ‘USA’ and ‘Hubei’ datasets are 1 hour and 15 minutes, respectively, the corresponding total forecast durations are 24 hours and 6 hours.
Notably, to evaluate the model's ability to learn precipitation evolution patterns across different spatial scales, the spatial resolution of ‘USA’ and ‘Hubei’ datasets was downsampled to 6km and 2km, respectively. In addition, the entire input MRMS QPE/radar domain was predicted for these two datasets. And considering the relatively small study area of the ‘Hubei’ dataset, its spatial coverage of the input ERA5 reanalysis data and satellite imagery is configured to exceed that of the predicted variable (detailed latitude and longitude ranges are specified in the spatial attribution maps provided in the supplementary materials).

Given the varying data ranges across different variables, each variable was normalized before being fed into the model--specifically, mean subtraction followed by division by the standard deviation, except for the MRMS QPE and radar data, which underwent min-max (0-1) normalization as required by the specific design of the loss function and for numerical stability. For these two datasets, each month's data is split sequentially into 80\% for training, 10\% for validation, and 10\% for testing. In addition, to enhance the generalization capability, we introduced random Gaussian noise (with a mean of 0 and a standard deviation of 0.02) to all normalized input variables during the training phase. This approach enables the model to exhibit strong robustness against outlier noise samples that may appear in the testing set.

% It should be noted that due to page limitations, only the experimental results on the ‘USA’ dataset are presented in the main text, while the results for the ‘Hubei’ dataset are provided in the supplementary materials.

\subsection{Experimental Details}
The training epoch is set to 200 with the lead time steps 2 (The number of iterations for the ‘LPM’ is also two). The initially learning rate is set to 1e-5. And we employ a warm-up strategy (20 epochs) followed by a cosine learning rate schedule. The AdamW optimizer is adopted with hyperparameters $\beta_1=0.9$ and $\beta_2=0.999$. Weight decay is uniformly set to 1. Additionally, dropout regularization is applied to the model, with rates of 0.15 and 0.2 for ‘USA’ and ‘Hubei’ datasets, respectively. All training process are run based on the PyTorch framework with 4 NVIDIA RTX 4090 GPUs.

\subsection{Compared Methods and Metric Indices}
To validate the proposed method, we select several leading methods for comparison, including ConvGRU \cite{convgru_cite}, SimVP \cite{Gao_2022_CVPR}, Metnet2 \cite{Espeholt2022} and EarthFarseer \cite{wu2024earthfarsser}. To enhance its predictive capability beyond the ultra-short-term time horizon, we incorporate the HTA algorithm \cite{bi2023accurate} into SimVP, referred to as SimVP\_HTA. It is noteworthy that these models represent the current SOTA in the field of precipitation forecasting. Specifically, the original Metnet models reported a forecasting horizon of 24 hours, while the original EarthFarseer paper achieved a maximum forecasting horizon of 10 hours. Moreover, both models were evaluated on precipitation datasets incorporating multi-source atmospheric observations in their respective studies and demonstrated excellent predictive performance. And the primary reason we selected some advanced nowcasting models (such as SimVP and ConvGRU) for comparison is the scarcity of existing publicly published research on 0–24 hour high-spatial-resolution precipitation forecasting.
These methods are reproduced on our devices and adhere to the original paper's hyperparameter settings as closely as possible.

This study employ several performance metrics to measure the consistency between the predicted variable and their truth, including Possibility of Detection (POD), Critical Success Index (CSI) and Heidke Skill Score (HSS) \cite{SIT2024106001}. The higher of POD, CSI and HSS indices indicate more accurate prediction results:
\begin{equation}
\label{pod}
\begin{aligned}
    \text{POD} = \frac{\text{TP}}{\text{TP} + \text{FN}}
\end{aligned}
\end{equation}
\begin{equation}
\label{csi}
\begin{aligned}
    \text{CSI} = \frac{\text{TP}}{\text{TP} + \text{FN} + \text{FP}}
\end{aligned}
\end{equation}
\begin{equation}
\label{hss}
\begin{aligned}
    \text{HSS} = \frac{\text{TP} \times \text{TN} - \text{FN} \times \text{FP}}{(\text{TP} + \text{FN}) \times (\text{FN} + \text{TN}) + (\text{TP} + \text{FP}) \times (\text{FP} + \text{TN})}
\end{aligned}
\end{equation}
\begin{equation}
\label{fbi}
\begin{aligned}
    \text{FBI} = \frac{\text{TP} + \text{FP}}{\text{TP} + \text{FN}}
\end{aligned}
\end{equation}
\noindent where TP (True Positives) indicates correctly forecasted events (e.g., predicted and observed precipitation), FP (False Positives) represents false alarms (predicted but not observed), FN (False Negatives) denotes missed events (observed but not predicted),and TN (True Negatives) reflects correct rejections (no event predicted and observed).

\subsection{Comparative Results on the ‘USA’ Dataset}\label{results_USA}  % 通过展示间隔4个小时的图来放大图以展示更多细节，把图例放在下面  ori:302  355  671/673  781/782    down: 355  671 673 679  781  782
\begin{figure*}[ht]
% \vskip 0.01in
\centering
    \includegraphics[width=0.5\textwidth,trim=0 0 0 0,clip]{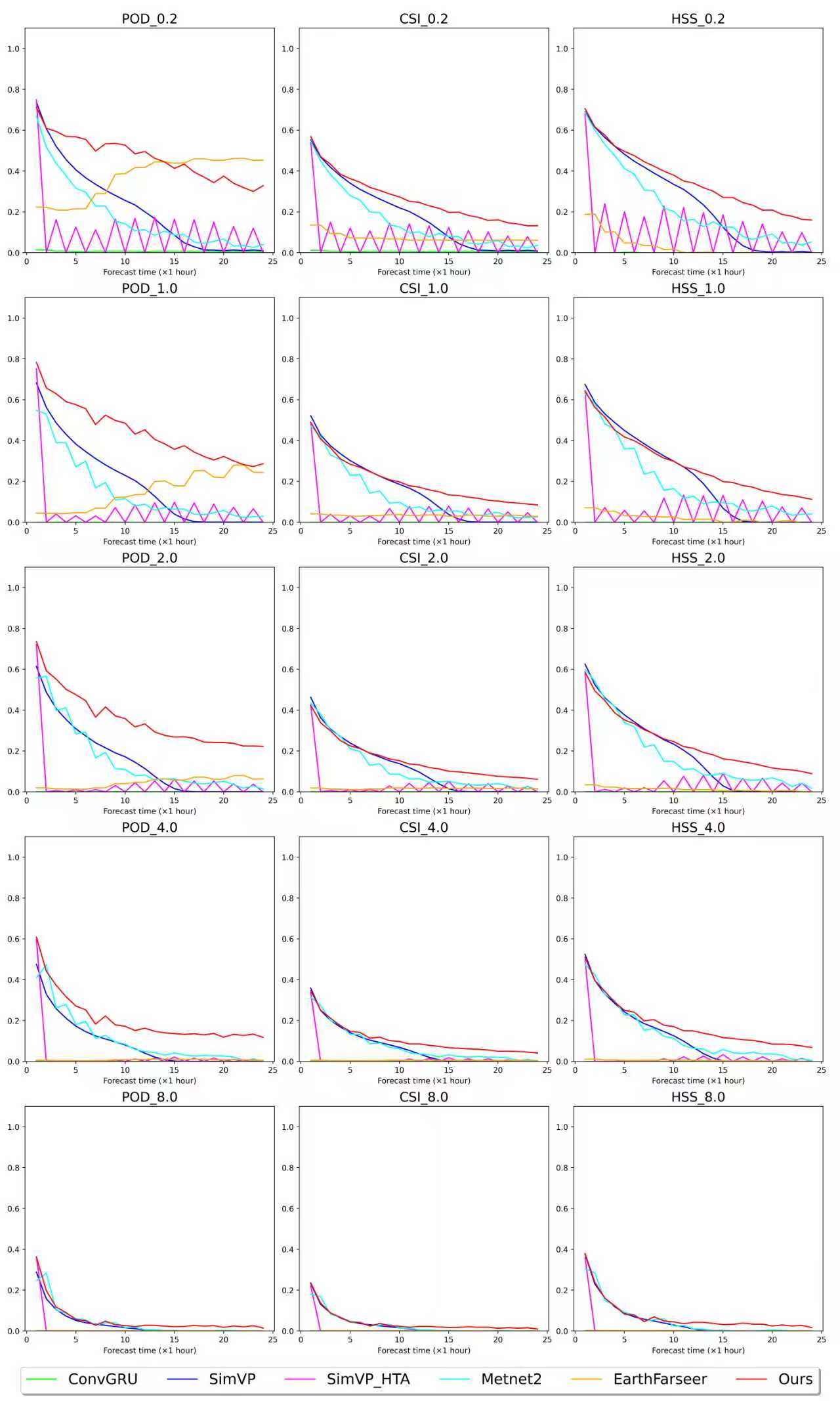}
    \caption{The 24 lead times' quantitative comparison results on ‘USA’ dataset. 0.2, 1, 2, 4 and 8mm/hour represent different Cumulative precipitation rate thresholds.}
    \label{fig:metric_nexrad}
% \vskip -0.2in
\end{figure*}
\begin{table}[htbp]
\centering
\caption{The 24 lead times' averaged quantitative comparison results on ‘USA’ dataset. The values 0.2, 2, and 8mm/h specify the precipitation rate thresholds used for calculating the metric indices, with the thresholds of 1 and 4 omitted from presentation here.}
\label{tab:performance}
\setlength{\tabcolsep}{2pt} % Reduce column spacing
\begin{tabular}{|l|ccc|ccc|ccc|}
\hline
Method & \multicolumn{3}{c|}{POD} & \multicolumn{3}{c|}{CSI} & \multicolumn{3}{c|}{HSS} \\ \hline
 & 0.2 & 2 & 8 & 0.2 & 2 & 8 & 0.2 & 2 & 8 \\ \hline
ConvGRU & 0.007 & 0.000 & 0.000 & 0.006 & 0.000 & 0.000 & -0.001 & 0.000 & 0.000 \\ \hline
SimVP & 0.218 & 0.153 & 0.035 & 0.181 & 0.119 & 0.029 & 0.254 & 0.184 & 0.052 \\ \hline
SimVP\_HTA & 0.099 & 0.047 & 0.015 & 0.077 & 0.032 & 0.009 & 0.110 & 0.048 & 0.015 \\ \hline
Metnet2 & 0.183 & 0.149 & 0.039 & 0.162 & 0.114 & 0.029 & 0.237 & 0.180 & 0.051 \\ \hline
EarthFarseer & 0.369 & 0.046 & 0.000 & 0.073 & 0.017 & 0.000 & 0.032 & 0.013 & 0.001 \\ \hline
Ours & \textbf{0.471} & \textbf{0.357} & \textbf{0.057} & \textbf{0.265} & \textbf{0.157} & \textbf{0.041} & \textbf{0.355} & \textbf{0.241} & \textbf{0.073} \\ \hline
% ConvGRU & 0.007 & 0.000 & 0.000 & 0.006 & 0.000 & 0.000 & -0.001 & 0.000 & 0.000 & \textcolor{red}{0.118} & \textcolor{red}{0.000} & \textcolor{red}{0.000} \\ \hline
% SimVP & 0.218 & 0.153 & 0.035 & 0.181 & 0.119 & 0.029 & 0.254 & 0.184 & 0.052 & \textcolor{red}{0.374} & \textcolor{red}{0.300} & \textcolor{red}{0.096} \\ \hline
% SimVP\_HTA & 0.099 & 0.047 & 0.015 & 0.077 & 0.032 & 0.009 & 0.110 & 0.048 & 0.015 & \textcolor{red}{0.242} & \textcolor{red}{0.156} & \textcolor{red}{0.063} \\ \hline
% Metnet2 & 0.183 & 0.149 & 0.039 & 0.162 & 0.114 & 0.029 & 0.237 & 0.180 & 0.051 & \textcolor{red}{0.293} & \textcolor{red}{\textbf{0.395}} & \textcolor{red}{0.204} \\ \hline
% EarthFarseer & 0.369 & 0.046 & 0.000 & 0.073 & 0.017 & 0.000 & 0.032 & 0.013 & 0.001 & \textcolor{red}{4.890} & \textcolor{red}{1.710} & \textcolor{red}{0.024} \\ \hline
% Ours & \textbf{0.471} & \textbf{0.357} & \textbf{0.057} & \textbf{0.265} & \textbf{0.157} & \textbf{0.041} & \textbf{0.355} & \textbf{0.241} & \textbf{0.073} & \textcolor{red}{\textbf{1.387}} & \textcolor{red}{1.889} & \textcolor{red}{\textbf{0.389}} \\ \hline
\end{tabular}
\end{table}

\begin{figure*}[ht] % 
\centering
    \includegraphics[width=1\textwidth,trim=0 0 0 0,clip]{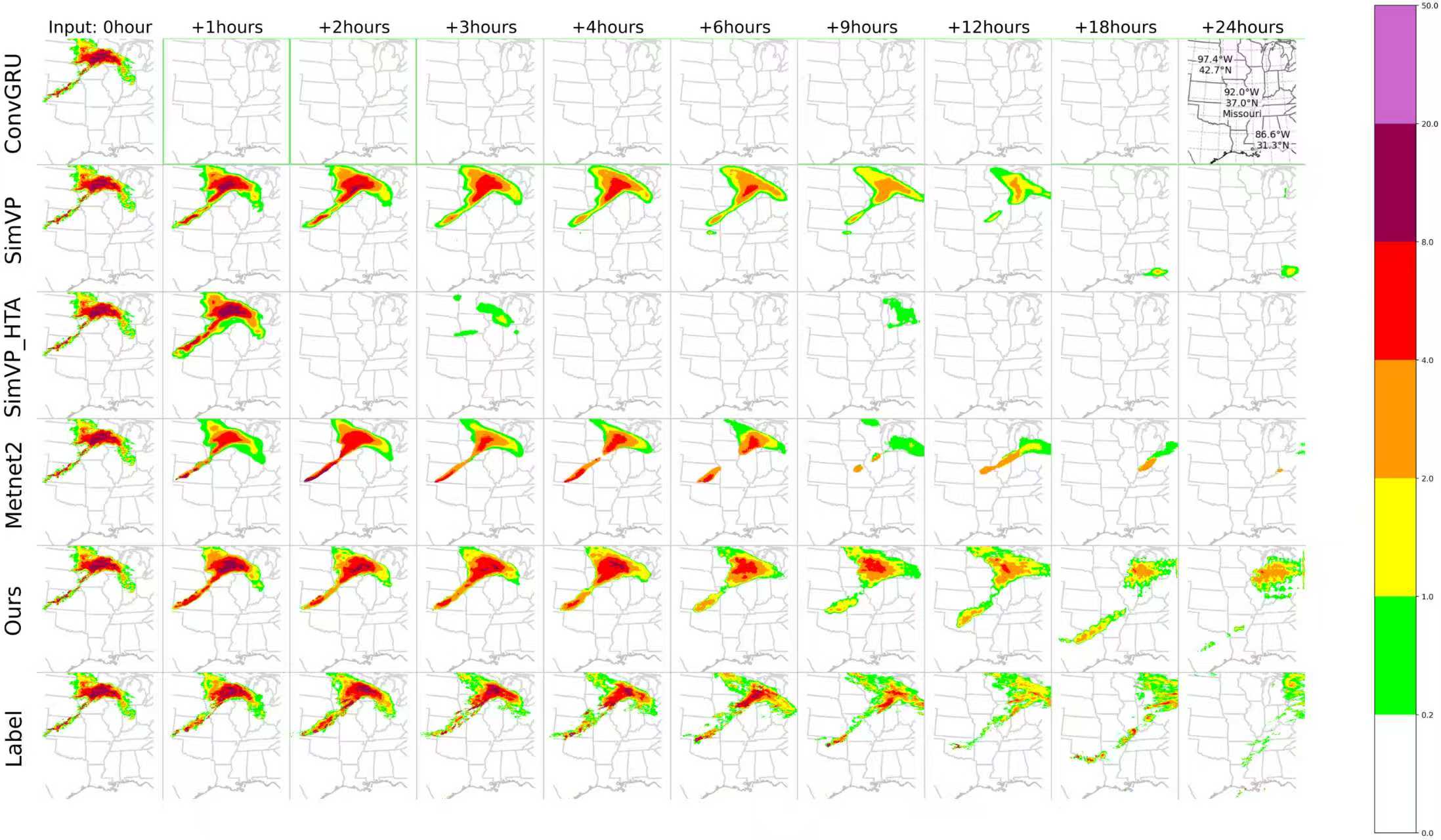} %20161026T030000
    \caption{The qualitative comparison on ‘USA’ dataset. The input data covers the period: 2016-10-26 03:00 - 2016-10-26-07:00 (UTC). The results of EarthFarseer are not shown here due to its poor qualitative performance.
}
    \label{fig:quali_fig_usa3}
\end{figure*}

\begin{figure*}[ht] % 
\centering
    \includegraphics[width=1\textwidth,trim=0 0 0 0,clip]{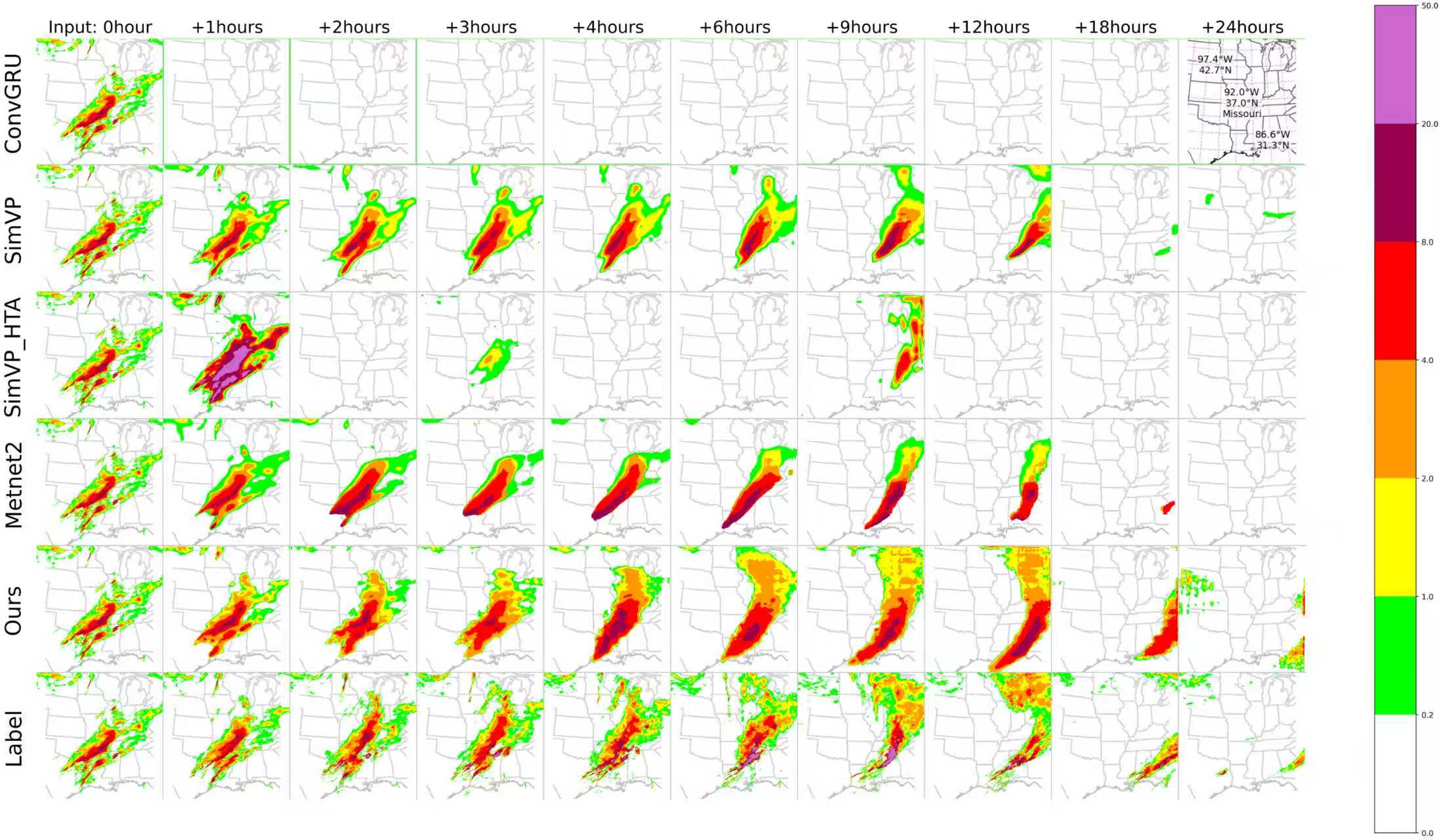}  % 20161128T120000
    \caption{The qualitative comparison on ‘USA’ dataset. The input data covers the period: 2016-11-28 12:00 - 16:00 (UTC).
}
    \label{fig:quali_fig_usa4}
\end{figure*}

\begin{figure*}[ht] % 
\centering
    \includegraphics[width=1\textwidth,trim=0 0 0 0,clip]{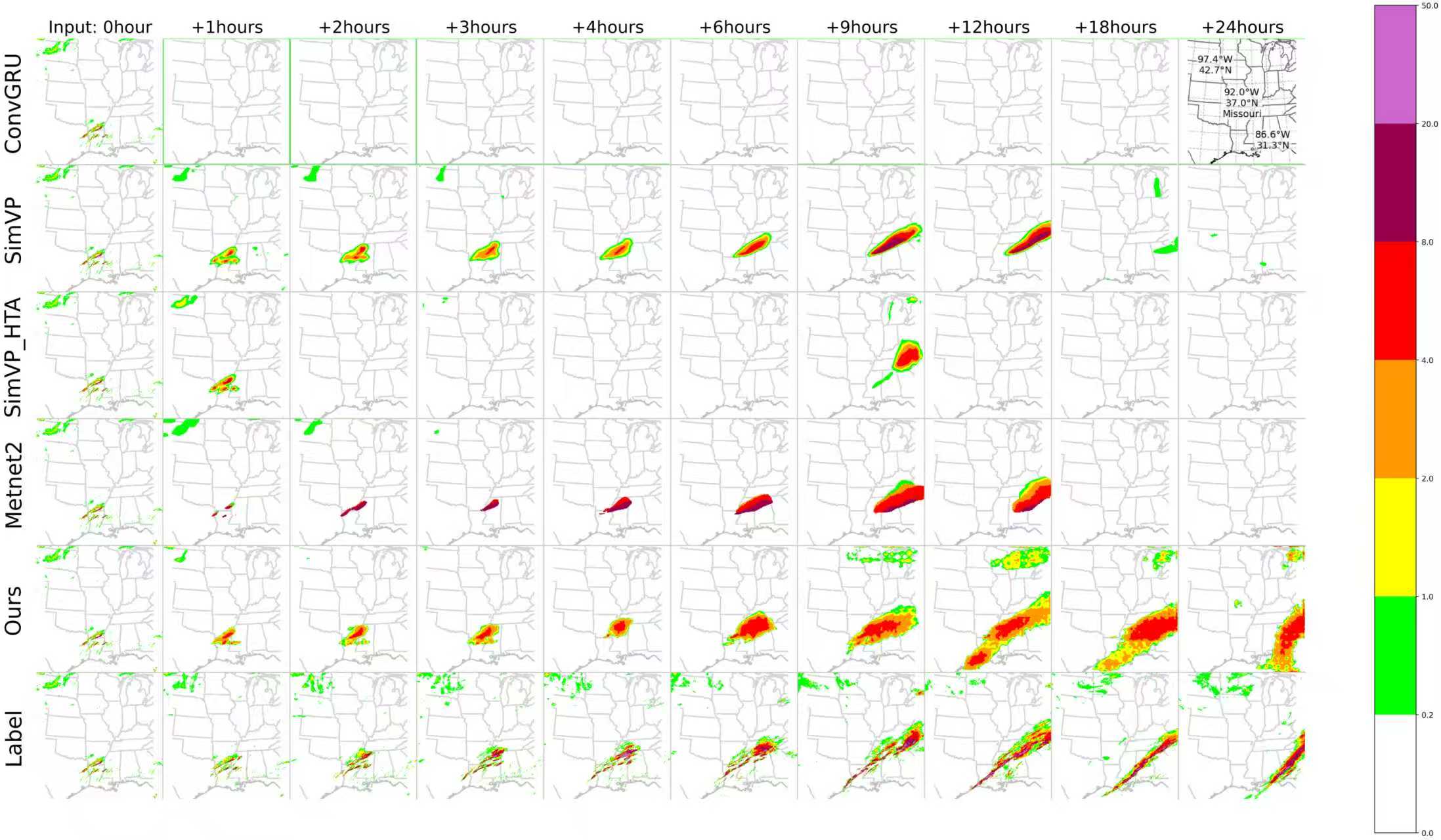} % 20161129T160000
    \caption{The qualitative comparison on ‘USA’ dataset. The input data covers the period: 2016-11-29 16:00 - 20:00 (UTC).
}
    \label{fig:quali_fig_usa5}
\end{figure*}
% Fig. \ref{fig:metric_nexrad} presents the quantitative comparison results of different methods on ‘USA’ dataset. It is evident that, our proposed method significantly outperforms all other methods in POD, CSI and HSS indices. 
%尽管我们方法在预测的前7到8个lead times性能劣于SimVP，但我们方法在中后期的这三个指标上表现出了最优的性能，而其他方法在中后期的指标都非常接近0.从表2中的平均指标也可以看出，我们方法的这三个指标都是最优。相比次优的方法，我们方法在三个指标上的性能提升平均幅度分别为109.209\%，21.397\%，27.609\%。尽管我们方法的FAR指标较高，表明了存在误检，但是ACC指标确非常接近1，表明我们方法的误检率并不高。
Fig. \ref{fig:metric_nexrad} presents the quantitative comparison results of different methods on the ‘USA’ dataset. It is evident that our proposed method significantly outperforms all other methods in POD, CSI, and HSS metrics. Although our method performs nearly on par with SimVP and MetNet-2 during the early prediction stages at the 1 mm/h and 2 mm/h thresholds, it demonstrates consistently superior predictive skill in the mid-to-late stages. In the mid-to-late stages, the performance advantage of our method over SimVP may be attributed to the fact that the latter was originally designed for 0--2 hour ultra-short-term precipitation nowcasting. Nevertheless, our method still outperforms the enhanced version of this approach, SimVP\_HTA.

As shown in Table \ref{tab:performance}, which presents the average performance metrics across 24 time steps, our method achieves the best scores across all three metrics. Compared to the second-best method, our approach shows average performance improvements of 60.556\%, 36.576\%, and 34.013\% on these three metrics, respectively. % 25.876,124.841,30.952///45.304,31.092,33.333///38.976,29.730,33.333
% 0.308333/0.154666
% 0.126666/0.094
% 0.203/0.147
% \begin{figure*}[ht] % 
% \centering
%     \includegraphics[width=1\textwidth,trim=0 0 0 0,clip]{picture/quali/nexrad/down/70_label_pred_all.png} % 20150329T120000
%     \caption{The qualitative comparison of 24 lead times on ‘USA’ dataset. The input data covers the period: 2015-03-29 12:00 - 16:00 (UTC).
% }
%     \label{fig:quali_fig_usa2}
% \end{figure*}

% Additionally, we present several qualitative comparison cases. As shown in Fig. \ref{fig:quali_fig_usa2} and \ref{fig:quali_fig_usa3}, 我们的方法都成功地预测出了降水事件的演变和消散过程，而其他方法则要么没有预测出降水带，要么预测降水带会快速消散。值得注意的是第一个例子中我们方法预测的降水带范围要比真值略大，这有待进一步提升。

% 此外，图Fig. \ref{fig:quali_fig_usa4} 和Fig. \ref{fig:quali_fig_usa5}展示了间隔为4个时刻的预测结果来查看更多的空间细节。如图Fig. \ref{fig:quali_fig_usa4}所示，降水带首先出现在了俄克拉荷马州和密苏里州及其附近，然后向东移动并且降水带范围逐渐变小。最终在约20-24小时后消散。我们的方法准确地预测出了降水带向东移动的趋势。并且准确地预测到了在8小时lead time时在阿肯色州的强降水中心，尽管预测的强度值略微低估。我们的模型还准确地预测出了在16-20小时lead times时，东北区域的降水带比东南区域的强降水带消散速度快的过程.相比之下，其他方法如SimVP和Metnet2只是准确地预测出了中心区域强降水带在前期的演变规律，但是SimVP方法预测强降水带的消散速度过快，而Metnet2则没有预测出东北区域的中等强度降水带。并且其他方法的预测结果遭受着严重的模糊效应，只能预测出大概的降水带轮廓，不能对细节进行刻画。而我们的方法则显现出了细节，尤其是在4小时和8小时lead times时展现出和真值较为一致的细节。

% Fig. \ref{fig:quali_fig_usa5}展示了输入数据包括2016-11-29 16:00-20:00(UTC)时刻的预测结果. 从预测结果可以看出，我们的方法准确地预测出了在4小时lead times时俄克拉荷马州东部和阿肯色州西部强降水事件的出现，并准确地预测出了该降水带向东移动和变长的趋势。尽管我们方法的预测降水区域比真值要宽，但是降水带的出现和演变趋势捕捉非常准确。相比之下，其他方法包括SimVP和Metnet2则只是预测出在4-12小时lead times降水带出现并向东移动的趋势，但在12小时后则错误地预测其会消散，并且在这个lead times阶段内对降水带范围的预测也偏小。上述定量和定性预测结果充分表明我们方法准确地捕捉到了降水事件发生和演变的规律。
% 整合下面两段
Additionally, we present several qualitative comparison cases with irregular time intervals. It is worth noting that we also present the precipitation rates at the last input time step, i.e., at 0 hour. As shown in Fig. \ref{fig:quali_fig_usa3}, our method successfully predicts the evolution and dissipation processes of precipitation events, while other methods all predict its dissipation more quickly to varying degrees.
Furthermore, as shown in Fig. \ref{fig:quali_fig_usa4}, our method accurately captures the eastward movement of the precipitation band and successfully predicts the heavy precipitation center over Arkansas at the 9-hour lead time, although the predicted intensity is slightly underestimated. Our model also correctly predicts that the precipitation band in the northeastern region dissipates more quickly than the heavy precipitation band in the southeastern region during the 12--18 hour lead times.

In comparison, other methods such as SimVP and Metnet2 only capture the early-stage evolution of the central heavy precipitation band. Moreover, SimVP predicts overly rapid dissipation of the heavy precipitation band, while Metnet2 fails to predict the moderate precipitation band in the northeastern region. Moreover, predictions from other methods suffer from severe blurring effects--only the general outline of the precipitation band is captured, with limited spatial detail. This may be because other methods only use a pixel-level MSE/MAE loss function. In contrast, our method produces finer details, especially at the 4-hour and 9-hour lead times, showing good consistency with the ground truth.

Fig. \ref{fig:quali_fig_usa5} shows prediction results based on input data from 16:00--20:00 UTC on November 29, 2016. The predictions show that our method accurately captures the formation process of heavy precipitation in eastern Oklahoma and western Arkansas, as well as its subsequent eastward movement and elongation. Although the predicted precipitation area of ours is wider than the ground truth, the occurrence and evolving trend of the precipitation band are well reproduced. In comparison, other methods including SimVP and Metnet2 only capture the initial eastward movement of the precipitation band at 1--12 hour lead times, but incorrectly predict its dissipation beyond 12 hours. In addition, throughout 6--12 hour lead times, the precipitation bands predicted by these methods are consistently smaller in spatial extent. 

The quantitative and qualitative results presented above strongly demonstrate that our method effectively captures the occurrence and evolution patterns of precipitation events. 
% Furthermore, the quantitative results on the Hubei dataset (Fig. S1 and Table S1) also indicate that our model achieves optimal predictive performance. The corresponding qualitative comparisons (Fig. S2, Fig. S3 and Fig. S4) consistently show that our method more accurately predicts the evolution of precipitation. This corroborates the robustness and capability of our proposed forecasting model, which delivers consistently strong performance across precipitation products with different modalities.

Note that we employed ERA5 reanalysis data, which has an inherent time latency, to validate the effectiveness of our forecasting model. This choice was made primarily due to its ease of access and low computational resource consumption. Additionally, we conducted forecast experiments using the hourly updated HRRR data \cite{ANorthAmerican}. As shown in Fig. S5, the quantitative comparison results demonstrate that our model also achieves high-precision forecasting performance. This exemplifies the feasibility and robustness required for the model's potential direct application in operational forecasting systems.

\subsection{Comparative Results on the ‘Hubei’ Dataset}\label{results_Hubei}  %  ori:707  54-60  157  929/934  down:98  930
% 图1展示了不同方法在‘Hubei’数据集上24个lead times（6小时）的定量比较结果。综合来看，我们的方法仍然取得了最优的性能，尤其是随着回波强度阈值的增加，和其他方法的预测精度差异越来越大。值得注意的是除了我们的方法，几乎所有其他方法的POD,CSI和HSS在阈值为30时的指标都非常接近0。此外，表3的平均性能指标也表明了我们方法的绝对优势。
% 我们还展示了在‘Hubei’数据集上的两个定性比较例子。如图1和2所示，这是两个间隔1小时（4个时刻）的雷达回波序列。我们的方法非常准确地预测了降水的出现时间，尤其是在第一个例子中。尽管演变规律细节并不完全准确，但是降水区域的增加趋势是一致的。相比之下，其他方法都没有预测到降水事件的发生。因此，我们的方法取得了最优的预测性能。
\begin{table}[ht]
\centering
\caption{The 24 lead times' averaged quantitative comparison results on ‘Hubei’ dataset. 20, 25, 30 dBZ represent the calculation of metric indices under different reflectivity thresholds. The results of EarthFarseer are not shown here due to its poor performance in this dataset.}
\label{tab:performance_hubei}
\setlength{\tabcolsep}{2pt}
\begin{tabular}{|l|ccc|ccc|ccc|}
\hline
Method & \multicolumn{3}{c|}{POD} & \multicolumn{3}{c|}{CSI} & \multicolumn{3}{c|}{HSS} \\ \hline
 & 20 & 25 & 30 & 20 & 25 & 30 & 20 & 25 & 30 \\ \hline
ConvGRU & 0.274 & 0.175 & 0.017 & 0.210 & 0.142 & 0.016 & 0.320 & 0.226 & 0.025 \\ \hline
SimVP & 0.323 & 0.252 & 0.106 & 0.195 & 0.152 & 0.074 & 0.294 & 0.240 & 0.126 \\ \hline
SimVP\_HTA & 0.254 & 0.188 & 0.050 & 0.191 & 0.142 & 0.042 & 0.292 & 0.227 & 0.071 \\ \hline
Metnet2 & 0.080 & 0.075 & 0.046 & 0.077 & 0.070 & 0.042 & 0.130 & 0.121 & 0.077 \\ \hline
Ours & \textbf{0.410} & \textbf{0.387} & \textbf{0.281} & \textbf{0.216} & \textbf{0.176} & \textbf{0.114} & \textbf{0.325} & \textbf{0.276} & \textbf{0.191} \\ \hline
\end{tabular}
\end{table}
\begin{figure*}[ht]
\centering
    \includegraphics[width=0.5\textwidth,trim=0 0 0 0,clip]{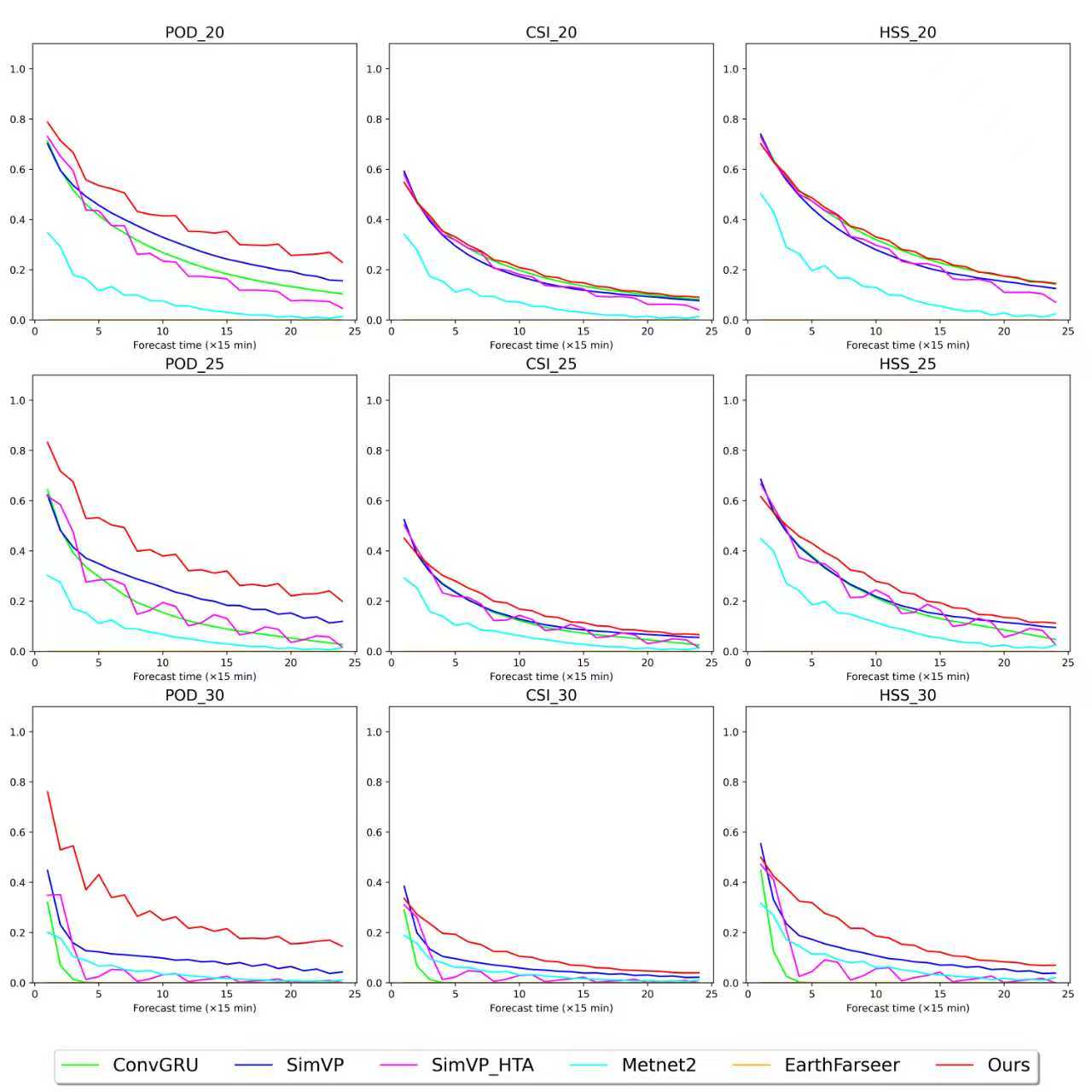}
    \caption{The 24 lead times' quantitative comparison results on ‘Hubei’ dataset. The values 20, 25, and 30 correspond to different radar reflectivity thresholds.}
    \label{fig:metric_Hubei}
\end{figure*}

\begin{figure*}[ht]
\centering
    \includegraphics[width=1\textwidth,trim=0 0 0 0,clip]{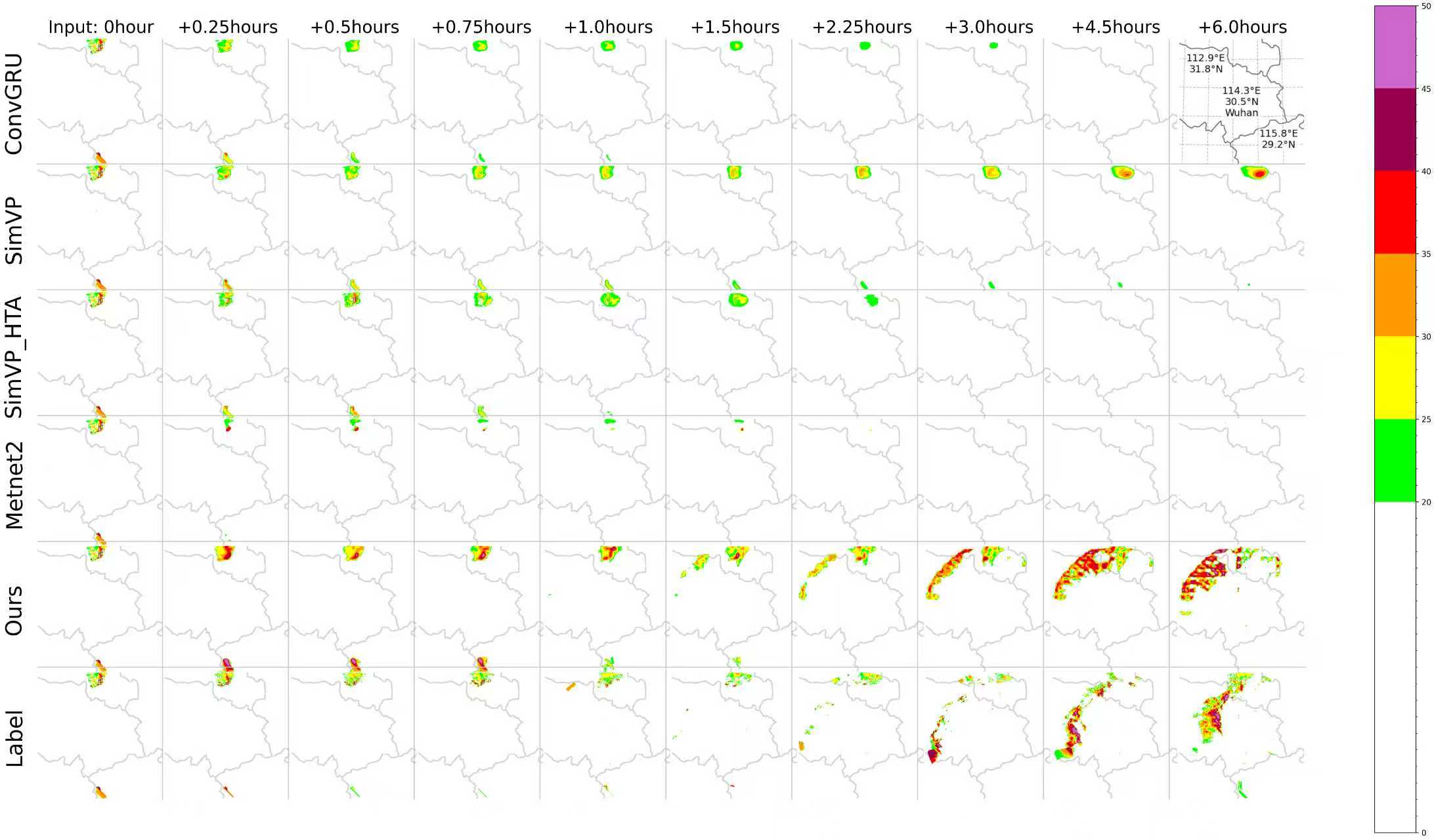} % 20220727150000
    \caption{The qualitative comparison on ‘Hubei’ dataset. The input data covers the period: 2022-07-27 15:00 - 16:00 (UTC).
}
    \label{fig:quali_fig_Hubei}
\end{figure*}
% \begin{figure*}[ht]
% \centering
%     \includegraphics[width=1\textwidth,trim=0 0 0 0,clip]{picture/quali/fy4b/original/60_label_pred_all.png} % 20220627153000
%     \caption{The qualitative comparison of 24 lead times on ‘Hubei’ dataset. The input data covers the period: 2022-06-27 15:30 - 16:30 (UTC).
% }
%     \label{fig:quali_fig_Hubei2}
% \end{figure*}
\begin{figure*}[ht]
\centering
    \includegraphics[width=1\textwidth,trim=0 0 0 0,clip]{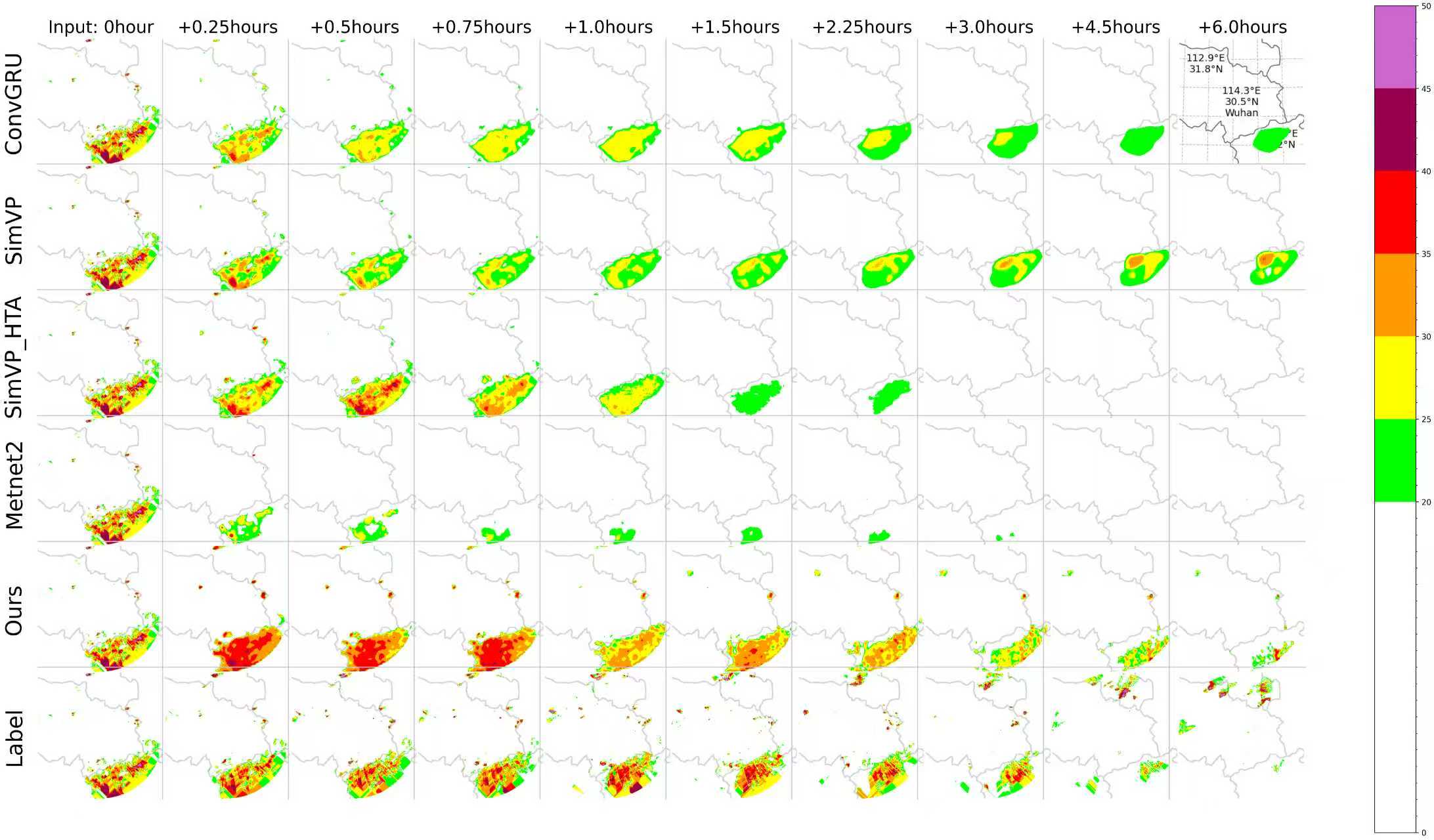} % 20220629053000
    \caption{The qualitative comparison on ‘Hubei’ dataset. The input data covers the period: 2022-06-29 05:30 - 06:30 (UTC).
}
    \label{fig:quali_fig_Hubei3}
\end{figure*}
\begin{figure*}[ht]
\centering
    \includegraphics[width=1\textwidth,trim=0 0 0 0,clip]{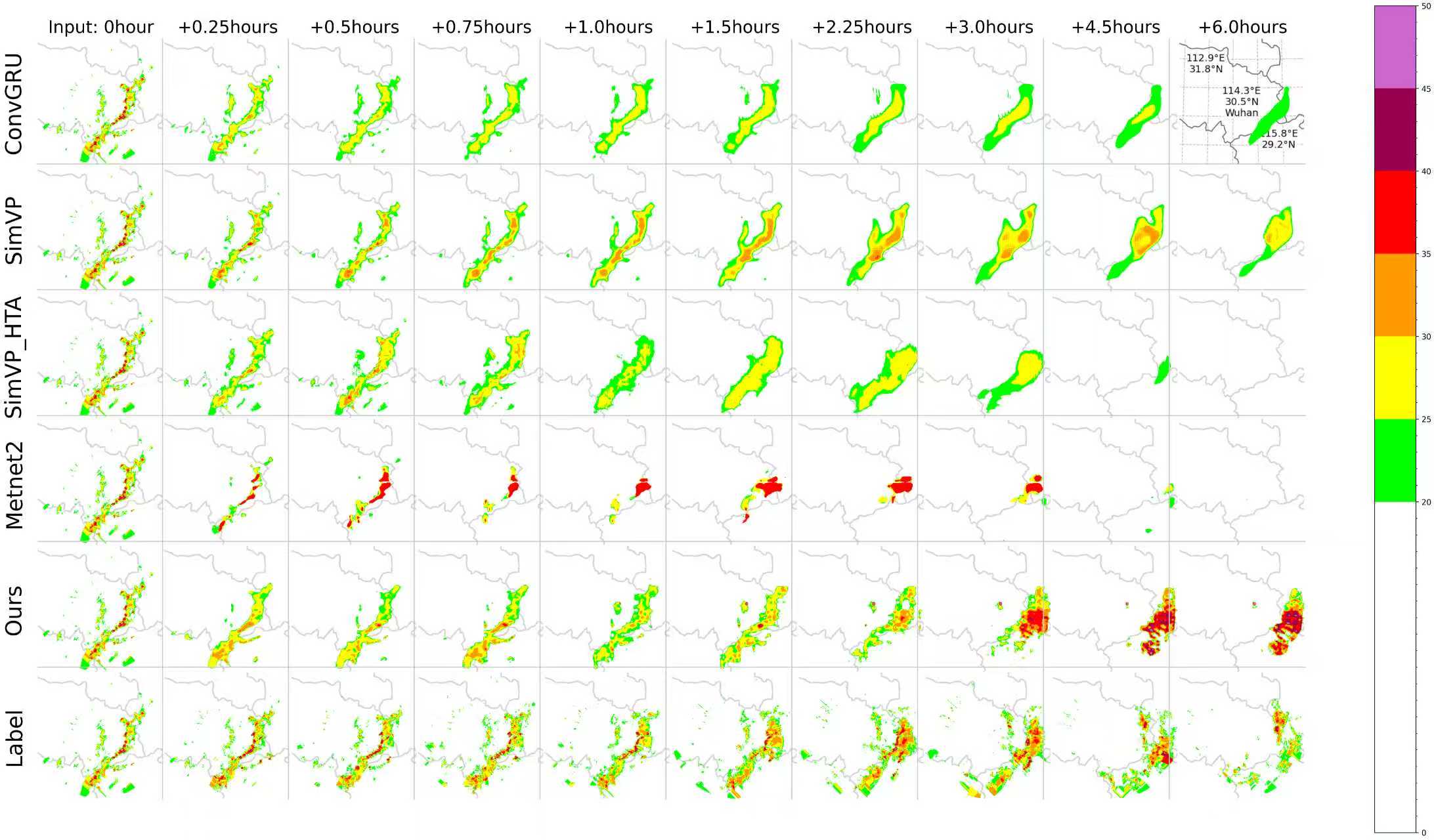}  % 20230827084500
    \caption{The qualitative comparison on ‘Hubei’ dataset. The input data covers the period: 2023-08-27 08:45 - 09:45 (UTC).
}
    \label{fig:quali_fig_Hubei4}
\end{figure*}

Fig. \ref{fig:metric_Hubei} presents the quantitative comparison of different methods across 24 lead times (6 hours) on the ‘Hubei’ dataset. Overall, our method consistently achieves superior performance, with the performance gap widening significantly as the reflectivity threshold increases. Furthermore, the average performance metrics in Table \ref{tab:performance_hubei} conclusively demonstrate our method's absolute advantage. Compared to the second-best method, our approach shows average performance improvements of 58.358\%, 20.428\%, and 20.151\% on three metrics (POD, CSI and HSS), respectively.
% We further present two qualitative comparison cases on 'Hubei' dataset. As demonstrated in Fig. \ref{fig:quali_fig_Hubei} and Fig. \ref{fig:quali_fig_Hubei2}, 我们方法非常准确地捕捉到了降水的出现时间和演变规律。尽管在Fig. \ref{fig:quali_fig_Hubei}中降水带提前半小时预测出来，但其他方法完全没有预测出西部降水带的出现。并且如Fig. \ref{fig:quali_fig_Hubei2}所示，我们方法成功地预测出强降水带将持续，而其他方法则预测出降水带将逐渐变小并消失。 

% 此外，我们还展示了间隔为4个time steps的预测结果。如图Fig. \ref{fig:quali_fig_Hubei3}最后一行所示，降水带区域和降水强度逐渐变小，并且在6小时lead times的时候完全消散。尽管我们的预测结果在5-6小时lead times的时候预测区域偏大，但是仍然比其他方法的预测精度高。其他方法的降水带预测结果要么基本没变化，要么消散太快，均和真值不符。图Fig. \ref{fig:quali_fig_Hubei4}还表明我们的方法准确地预测出了降水强度的增大和降水带的东移，而其他方法则存在强度值低估现象，并且他们的降水区域预测结果基本不变或变化太快。
% 整合下面两段
We further present several qualitative comparison cases on the ‘Hubei’ dataset. As demonstrated in Fig. \ref{fig:quali_fig_Hubei}, our method accurately captures the timing of precipitation, while other methods completely fail to predict the emergence of the western precipitation band.
Additionally, as shown in the last row of Fig. \ref{fig:quali_fig_Hubei3}, the precipitation area and intensity gradually diminish, completely dissipating by the 6-hour lead time. Although our predictions slightly overestimate the coverage, they still accurately capture the trend of precipitation dissipation. The precipitation bands predicted by other methods either remain largely unchanged or dissipate too quickly, both of which deviate from the ground truth. Fig. \ref{fig:quali_fig_Hubei4} further demonstrates that our method accurately predicts the increase in precipitation intensity and the eastward movement of the precipitation band, whereas other methods exhibit underestimation of intensity values and their predicted precipitation areas either remain almost static or change too rapidly.

It can be observed that Metnet2 performs worse on the ‘Hubei’ dataset compared to the ‘USA’ dataset. This may be due to the fact that Metnet2 was originally designed for forecasting precipitation rate (mm/h) (as in the ‘USA’ dataset), and was not intended for predicting radar reflectivity (dBZ) as in the ‘Hubei’ dataset. Although precipitation rate (mm/h) and radar reflectivity (dBZ) are strongly correlated, they are related through a nonlinear conversion formula (the Z-R relationship), which may introduce significant differences in value ranges and data distributions. Additionally, the forecast interval divisions used by Metnet2 are inconsistent between these two datasets. Specifically, the ‘USA’ dataset includes 6 intervals: [0–0.2], [0.2–1], [1–2], [2–4], [4–8], and [$>$8]; while the ‘Hubei’ dataset includes 4 intervals: [0–20], [20–25], [25–30], and [$>$30]. We opted for these intervals instead of the dense interval divisions from the original study primarily to reduce the complexity of the learning task. Therefore, the combination of differences in numerical ranges and data distributions due to the nonlinear Z-R conversion, along with the substantially distinct interval divisions, explains why Metnet2 performs worse over ‘Hubei’ dataset than the ‘USA’ dataset. This further underscores the robustness and capability of our proposed precipitation forecasting model, which consistently delivers excellent predictive performance across precipitation products with different modalities.

Above all quantitative and qualitative results on two datasets corroborate the robustness and capability of our proposed forecasting model, which delivers consistently strong performance across precipitation products with different modalities.

\subsection{Ablation Study}
\subsubsection{The Effectiveness of the Designed Loss Functions}
\begin{figure*}[ht]
\centering
    \includegraphics[width=0.5\textwidth,trim=0 0 0 0,clip]{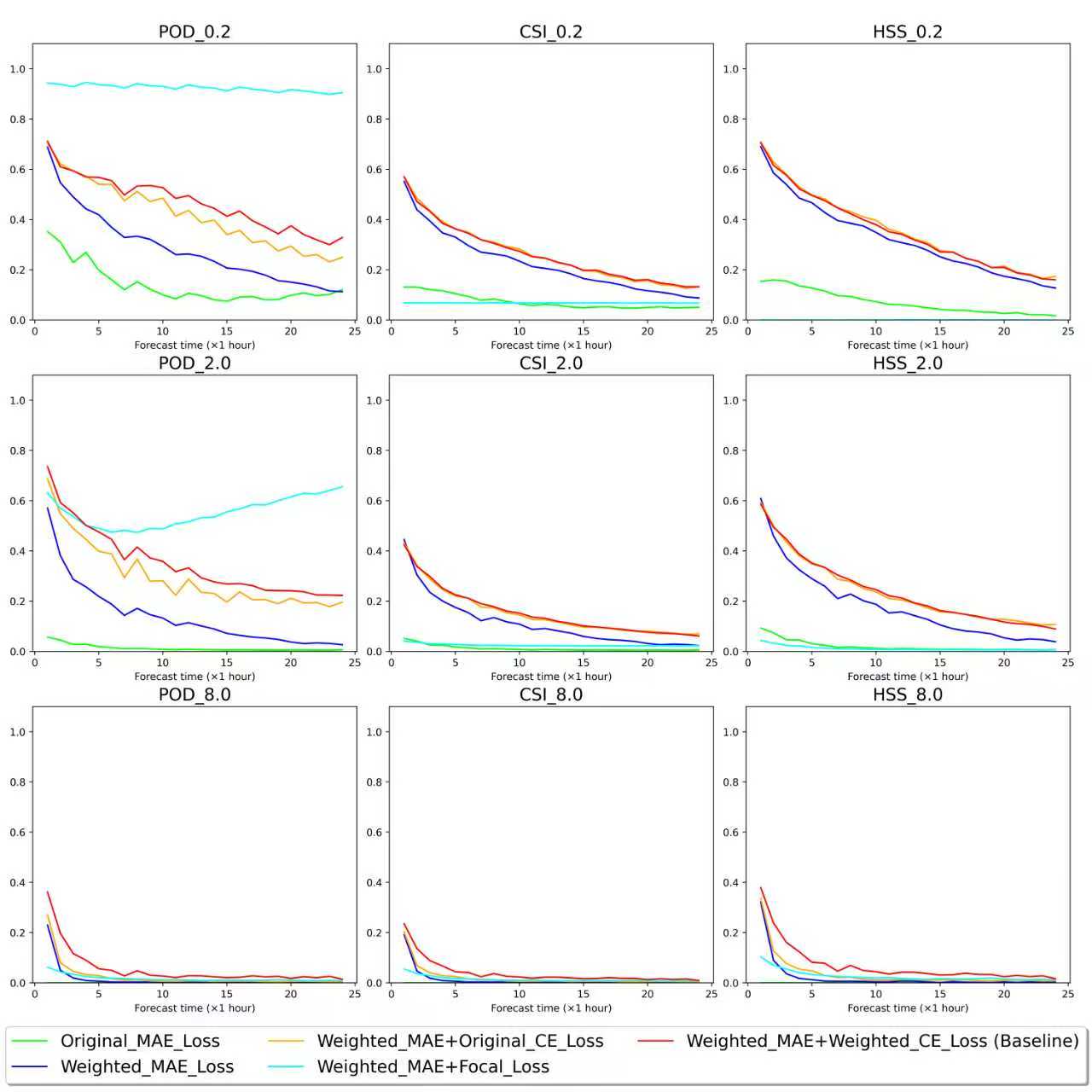}
    \caption{The quantitative ablation comparison results of loss functions on ‘USA’ dataset. Note that ‘Weighted\_MAE\_Loss’ refers to the MAE loss weighted by the parameters defined in Eq. \ref{a9990} (The weighting schemes are identical across different ablation studies). The term ‘Weighted\_MAE+Original\_CE\_Loss’ denotes a combination of the weighted MAE loss (with non-precipitation samples masked out, as in Eq. \ref{a9990}) and the unweighted cross-entropy loss. The focal loss is adopted from \cite{Lin_2017_ICCV}.
    }
    \label{fig:metric_loss}
\end{figure*}
\begin{figure*}[ht]
\centering
    \includegraphics[width=0.5\textwidth,trim=0 0 0 0,clip]{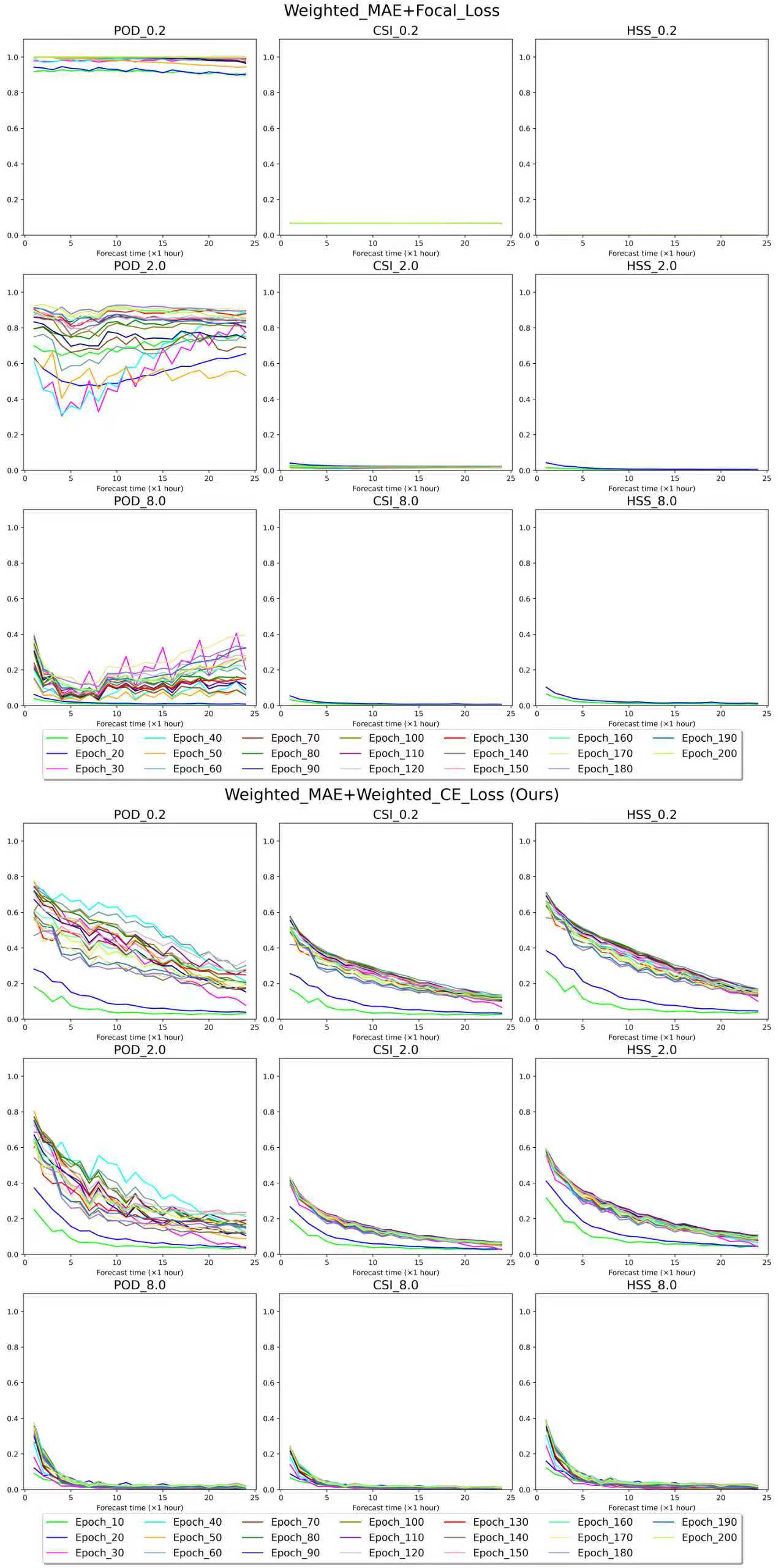}
    \caption{The quantitative ablation comparison results between ‘Weighted MAE+Focal’ loss and ‘WMCE’ loss on ‘USA’ dataset across different epochs. 
    % Pixel+其他loss实验中的Pixel loss只针对有雨样本。
    }
    \label{fig:metric_focal_elt}
\end{figure*}
% 为了验证所提出的损失函数的有效性，我们比较了不同的损失函数组合。从图1的绿线和蓝线可以看出原始的和加权的像素级MAE loss都难以检测到降水事件的发生，其大部分指标都非常接近0。这主要是由于像素级loss只关注高概率事件，且对漏检事件的惩罚较低。加入交叉熵loss后，如紫线所示，性能有所提升。这主要是由于交叉熵loss中的log函数对于漏检样本的惩罚会更大。但由于对无雨样本漏检的惩罚也大所以对高概率事件的过度关注问题并没有解决，性能并不突出。
%我们还试验了Pixel+Focal loss的组合，但是其误检率太高。这是由于Focal loss包含两个待调节的超参数，对于不同的数据集都要进行多次实验性地调整。并且和CE loss一样，Focal loss对于降水事件loss一直大于0，导致反射率值被高估。综合来看，我们的方法取得了最优的性能，在POD,CSI和HSS指标上均为最优并且没有需要调节的超参数。并且在FAR和ACC指标上和Pixel+ce loss持平。这充分证明了我们提出的损失函数在预测服从极端长尾分布的降水事件时的有效性。
To mitigate the issue where traditional MAE loss functions overly focus on a large number of non-precipitation samples, this study constructs a weighted cross-entropy loss function to improve the predictive capability for extremely scarce precipitation samples. Additionally, by combining it with a weighted MAE loss function, a ‘WMCE’ loss function is designed to accurately constrain the predicted precipitation intensity values. To validate the effectiveness of the designed loss functions, especially the ‘WMCE’ loss, we compared different loss function combinations in this part.

As shown by the green line in Fig. \ref{fig:metric_loss}, the original pixel-wise MAE loss fail to detect precipitation events effectively, with most metrics close to zero. This primarily occurs because original MAE loss focuses solely on high-probability events while imposing weak penalties for missed detections. After incorporating the logarithmic weighting of the original precipitation rate values, as shown by the blue line in Fig. \ref{fig:metric_loss}, the detection capability for precipitation events has improved. This is because higher precipitation rates receive greater weights, leading to a larger loss penalty for missed detections.

After incorporating the CE loss (orange line in Fig. \ref{fig:metric_loss}), we observe a further improvement in performance. This enhancement stems from the logarithmic function in the CE loss, which imposes stronger penalties for missed detection samples. However, since it also imposes substantial penalties on false alarms, the fundamental issue of excessive focus on high-probability events remains unresolved, resulting in sub-optimal overall performance. Accordingly, we introduced a weighted cross-entropy loss (as in Eq. \ref{a9991}) to place greater emphasis on rare samples, i.e., heavy precipitation events. As indicated by the red line in Fig. \ref{fig:metric_loss}, the results demonstrate a notable improvement in the POD compared to the original unweighted cross-entropy loss. In particular, for precipitation events exceeding 8 mm/h, all three metrics—POD, CSI, and HSS—show clear enhancement.

We also evaluated the predictive capability of focal loss \cite{Lin_2017_ICCV} for precipitation events with the unbalanced distribution by testing a combined ‘Weighted\_MAE + Focal\_loss’ approach, as illustrated in Fig. \ref{fig:metric_loss}. The parameters of the focal loss were set to their default values: $\alpha = 0.25$ and $\gamma = 2.0$. As shown by cyan line in Fig. \ref{fig:metric_loss}, this configuration yields a very high POD but near-zero CSI and HSS, indicating that the model tends to predict persistent precipitation with a high false alarm rate. Furthermore, as shown in Fig. \ref{fig:metric_focal_elt}, the performance of ‘Weighted\_Pixel + Focal\_loss’ across different training stages reveals significant instability, with large fluctuations in the POD metric. In contrast, the performance of our proposed ‘Weighted\_Pixel + ‘Weighted\_CE\_loss’ increases steadily as training progresses. The instability and ineffectiveness of focal loss stems from that it contains two tunable hyperparameters that require extensive experimental calibration for different datasets, while our designed ‘WMCE’ loss function does not involve any tunable hyperparameters. Overall, as shown in Fig. \ref{fig:metric_loss}, comprehensive evaluation demonstrates that our method achieves state-of-the-art performance, attaining optimal scores across all metrics (POD, CSI, and HSS). These results conclusively validate the effectiveness of our designed loss functions in predicting precipitation events following unbalanced distribution, addressing key limitations of existing approaches.

\subsubsection{Latent Space Iteration and Additional Atmospheric Variables} % 查看预测不同时刻对输入的归因的差异

\begin{figure}[t]
% \vskip 0.01in
\centering
\includegraphics[width=0.5\textwidth,trim=0 0 0 0,clip]{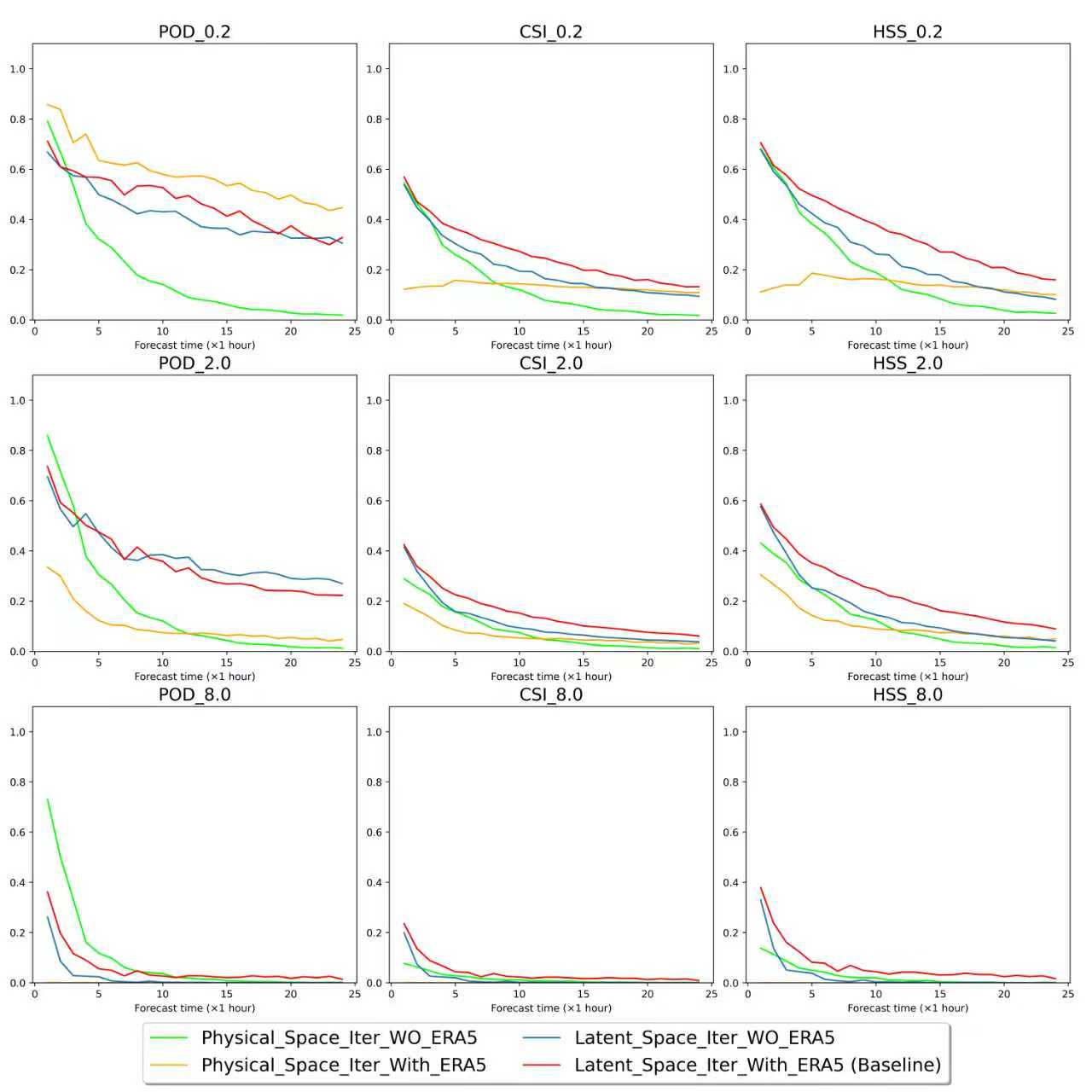}
    \caption{The ablation results of latent space iterative prediction and inputting more variables (ERA5) on ‘USA’ dataset. 
    % The displayed variables include U10m, mslp, Z500 and T850. ‘baseline’ means perform the iterative prediction in the latent space with 32 latent feature channels and input much more variables.
    ‘Physical\_Space\_Iter\_WO\_ERA5’ denotes the iterative prediction conducted in the physical variable space, employing only MRMS QPE and radar data while excluding ERA5 variables.‘Latent\_Space\_Iter\_With\_ERA5’ means perform the iterative prediction in latent variable space with inputting more variables (including MRMS QPE, radar and ERA5 variables). 
    % And ‘HTA’ algorithm is incorporated in all above experiments.
    }
    \label{fig:latent_iter}
% \vskip -0.2in
\end{figure}
To accurately predict the occurrence and evolution of precipitation, it is necessary to input extensive observations to accurately simulate the evolution of atmospheric states. Nevertheless, iterative forecasting in physical space requires learning the evolution patterns of all input physical variables, which is highly challenging. In contrast, iterative forecasting in the latent space only requires learning the evolution patterns of latent features strongly associated with precipitation. Therefore, in this part, we validated the effectiveness of iterative forecasting in the latent space and the inclusion of additional atmospheric observations (ERA5).

As shown by the green line in Fig. \ref{fig:latent_iter}, when only radar and MRMS QPE data are input (without ERA5 reanalysis data), the performance of iterative forecasting in physical space is very poor, with POD, CSI, and HSS metrics deteriorating rapidly within just a few hours of lead time. When ERA5 variables are added, as indicated by the orange line in Fig. \ref{fig:latent_iter}, the complexity of the learning task increases significantly, leading to a substantial decline in forecast performance at early stages, although the fluctuation in performance metrics is reduced.

In comparison, the performance of iterative forecasting in the latent space is far superior to that in physical space. Moreover, forecast performance is further enhanced with the inclusion of more atmospheric variables from ERA5. As demonstrated by the red line in Fig. \ref{fig:latent_iter}, performance across all lead times is enhanced, indicating that the model more accurately captures the evolution patterns of precipitation.
%我们还验证了在隐空间迭代预测和加入更多大气观测（ERA5）的有效性。为了准确预测降水的出现和演变，我们需要输入大量观测来准确地模拟大气状态的演变。在物理空间迭代预测需要学习所有物理变量的演变规律，这是非常困难的。而在隐空间迭代预测只需要学习和降水相关的隐特征的演变规律。如图1种绿线所示，当仅输入雷达和MRMS QPE数据时（不输入ERA5再分析数据），在物理空间迭代预测的性能非常差，POD,CSI和HSS指标在几个小时的预测时间内就会迅速恶化.而当加入ERA5变量后，如图1橙线所示，学习任务的复杂性大大增加，预测性能在前期就大幅下降，尽管性能指标的波动程度减小。而在隐空间迭代预测的性能要远优于物理空间迭代预测。并且随着输入大气观测变量的增加，预测性能进一步提升。如图中红线所示，各个lead times的性能都得到了提升，表明模型更加准确地捕捉到了大气状态的演变规律。
These experimental findings conclusively demonstrate the fundamental advantage of latent-space iterative prediction and the critical importance of incorporating additional atmospheric variables.

\subsubsection{Importance Sampling} % 查看预测不同时刻对输入的归因的差异
\begin{figure}[t]
% \vskip 0.01in
\centering
\includegraphics[width=0.5\textwidth,trim=0 0 0 0,clip]{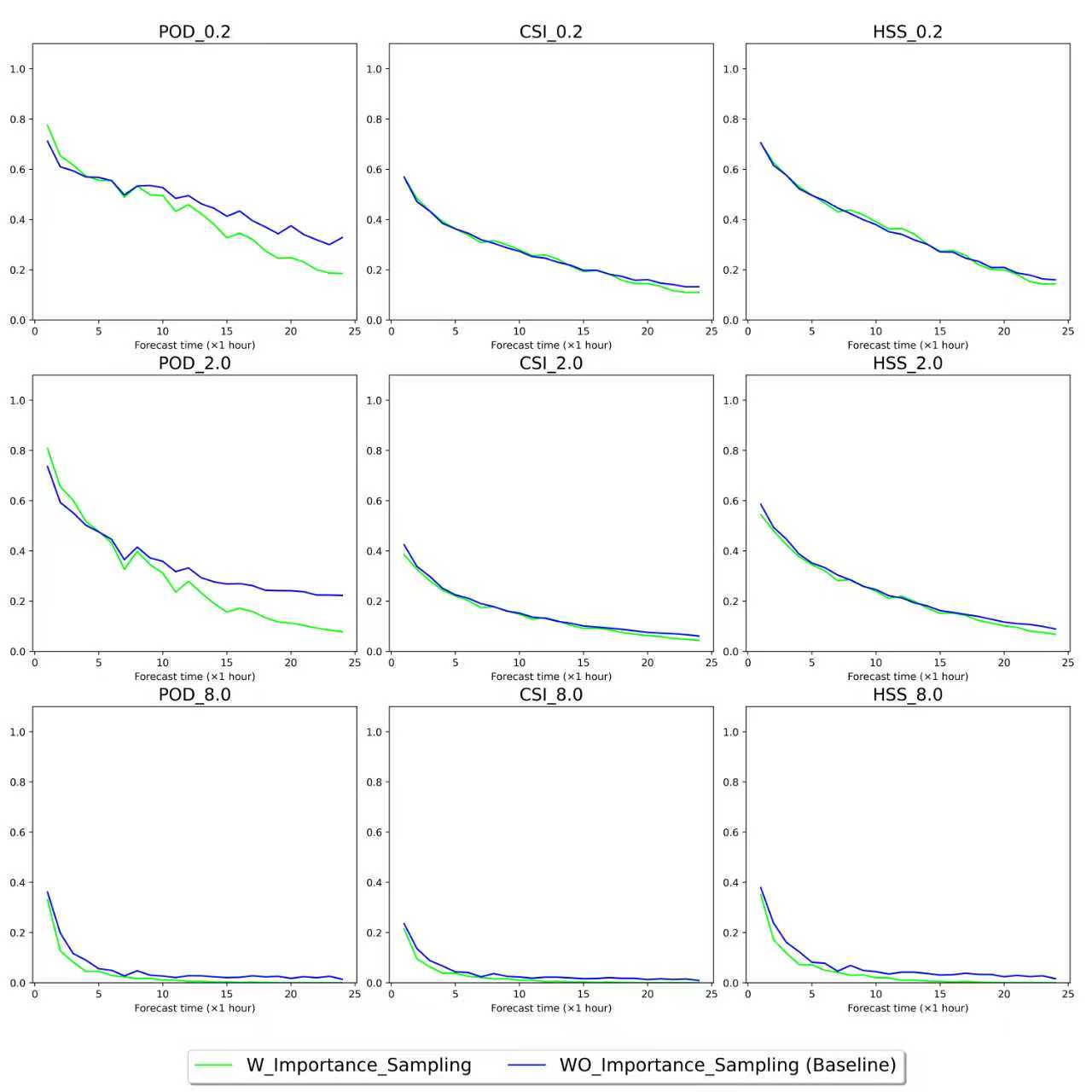}
    \caption{The ablation results of ‘Importance Sampling’ on ‘USA’ dataset. 
    % The displayed variables include U10m, mslp, Z500 and T850. ‘baseline’ means perform the iterative prediction in the latent space with 32 latent feature channels and input much more variables.
    ‘W\_Importance\_Sampling’ refers to the incorporation of importance sampling during the training phase to preprocess the training data by filtering out a portion of non-precipitation sequences. ‘WO\_Importance\_Sampling’, on the other hand, does not employ this technique and uses all available data for training.
    % And ‘HTA’ algorithm is incorporated in all above experiments.
    }
    \label{fig:Importance_Sampling}
% \vskip -0.2in
\end{figure}
Some studies have adopted the 'importance sampling' technique, which reduces the number of non-precipitation samples in the training set to allow models to focus more on learning the patterns of precipitation evolution. In this part, we evaluate the impact of the ‘importance sampling’ \cite{Espeholt2022} data preprocessing technique on model performance. As shown in Fig. \ref{fig:Importance_Sampling}, after incorporating the importance sampling technique, the POD metric decreases significantly in the mid-to-late stages, indicating a decline in the model’s ability to detect precipitation events. In addition, both the CSI and HSS metrics show a slight drop during the later period. Furthermore, the prediction accuracy for precipitation events exceeding 8 mm/h also deteriorates, as reflected by the worsening of the three evaluation metrics. This may be because non-precipitation data could potentially serve as precursor signals for subsequent precipitation events. For instance, the absence of precipitation might indicate current atmospheric conditions with low humidity but high temperature, which could later lead to precipitation through rapid evaporation of surface moisture. If such non-precipitation data are removed, the training dataset contains fewer potential signals that precede precipitation events, thereby reducing the model's capability to detect them.
% 在这一小节我们测试了‘重要性采样’数据预处理技术对于模型性能的影响。如图所示，当加入‘重要性采样’技术后，POD指标有了一定幅度下降，即模型对降水事件的检测能力有所下降。并且CSI和HSS指标在后期略微下降。这可能是由于无降水的数据可能是后续降水事件发生的潜在信号，比如无降水可能表示当前大气环境是低湿度但是高温度，后续随着地表水汽快速蒸发可能会形成降水。而如果将无降水数据去掉后，训练数据中导致降水事件发生的潜在信号变少，因此模型对降水事件的检测能力下降。

\subsection{Input Variables Contribution Analysis}
\begin{figure*}[t]
\centering
    \includegraphics[width=0.5\textwidth,trim=0 0 0 0,clip]{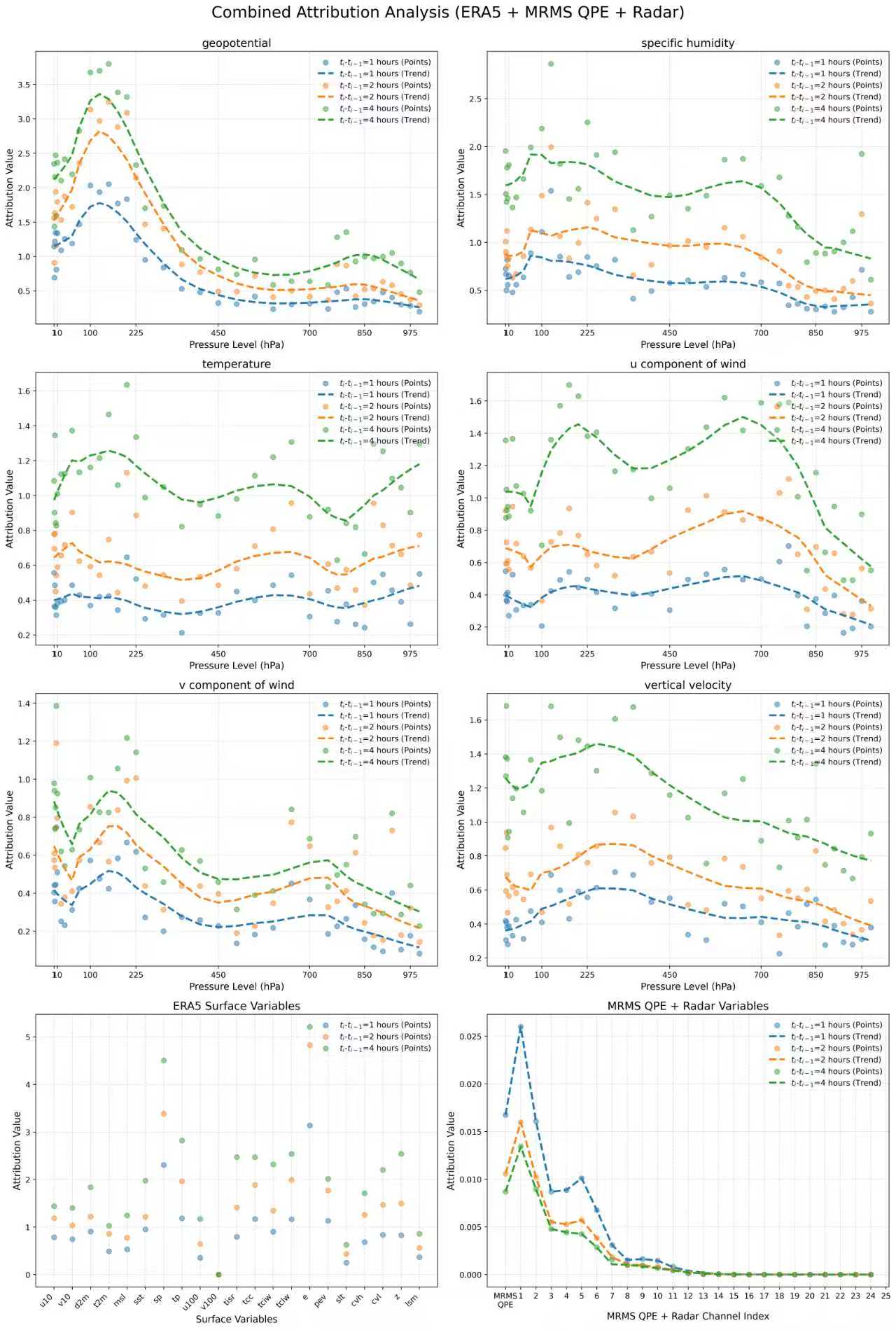}
    \caption{Attribution magnitude of input variables on ‘USA’ dataset. The dots and lines in different colors represent different prediction intervals. Please refer to the supplementary materials for more information on ERA5 surface variables.
    }
    \label{fig:grad_usa}
\end{figure*}
% Ablation experiments demonstrate that incorporating additional atmosphere variables into the input could improve the prediction performance. In order to visualize the contribution of multiple input variables, we utilize the 'Integrated Gradients' method \cite{pmlr-v70-sundararajan17a} to attribute predictions to the input variables, as shown in Fig. \ref{fig:grad_usa} and \ref{fig:grad_hubei}. 可以看出，‘USA’数据的输入大气变量中高海拔等级的变量比低海拔等级变量相对具有更大的归因值（比如‘geopotential’,‘u/v component of wind’ and‘vertical velocity’）。而‘Hubei’数据集的可视化结果表明低海拔等级的变量具有相对更大的归因值（比如‘geopotential’, ‘temperature’ and ‘u/v component of wind’）。这是由于‘USA’数据集和‘Hubei’数据集的地理位置和预报周期不一样导致的。USA数据集位于中纬度（纬度范围29.6°N-44.6°N），其天气变化主要由西风带中的高空槽脊系统驱动。其中位势高度直接定义高空槽（低压）和脊（高压）的位置与强度，而风场（U/V Wind）提供了涡度平流和风切变，垂直速度（Vertical Velocity）则直接表征大尺度上升运动。相比之下Hubei数据集位于低纬度（纬度范围28.3°N-32.7°N），天气更受低层热力与水汽的主导，其中温度决定了大气的静力稳定度和对流能量，而风场（U/V Wind）负责水汽输送和低空辐合。
% 另一方面对于不同的分辨率和预报时间/周期，起主导作用的大气变量也不同。低分辨率长周期预测（USA数据集的空间分辨率为6km，预测间隔和周期分别为1小时/24小时）关注大尺度天气形势的演变。高空的垂直速度（Vertical Velocity）变量能指示大范围、持续的上升运动区。而高空的U/V Wind和Geopotential则定义了大尺度动力过程的框架。高分辨率短周期预测（‘Hubei’数据集的空间分辨率为2km，预测间隔和周期为0.25和6小时）主要预测降水的具体发生、强度和消散，温度（Temperature）等低层变量直接反映了局地的能量状况和不稳定性。
% 此外，对于‘Hubei’数据集的输入数据中的FY4B卫星影像，前六个波段的归因值相对更高。这主要是由于FY4B前六个波段主要包含能够识别云相态、云滴大小等的可见光和近红外波段。其中可见光波段（如波段1、2） 能清晰呈现云团的纹理、边界和形态特征。而波段3（0.75-0.90μm） 和波段4（1.371-1.386μm） 对水汽敏感。近红外波段（特别是波段5、6）则对云中粒子的相态（水滴或冰晶）和大小非常敏感。最后，对于雷达数据，近地面的雷达观测归因最大，并且随着海拔的增加归因迅速变小，由于我们预测了近地面雷达反射率，这是非常合理的。
% 模型的归因结果证明，其推理方式与气象学原理高度一致，能够根据地域和任务需求，自适应地聚焦于主导其物理过程的关键变量，表明其较准确地学习到了隐藏在数据背后的大气运动规律。
Ablation studies demonstrate that incorporating additional atmospheric variables into the input improves prediction performance. To visualize the contribution of multiple input variables, we employ the ‘Integrated Gradients’ method \cite{pmlr-v70-sundararajan17a} to attribute predictions to the input variables, as shown in Fig. \ref{fig:grad_usa}.  

It can be observed that for the ‘USA’ dataset, higher-altitude atmospheric variables exhibit relatively larger attribution values compared to lower-altitude variables (e.g., ‘geopotential’, ‘u/v component of wind’ and ‘vertical velocity’). The primary reasons lie in the fact that, on the one hand, the ‘USA’ dataset is located in the mid-latitudes (27.6°N--46.4°N), where weather variability is primarily driven by upper-level trough and ridge systems within the westerlies. Here, ‘geopotential’ height directly defines the position and intensity of upper-level troughs (low pressure) and ridges (high pressure), while the wind field (‘u/v component of wind’) provides vorticity advection and wind shear, and ‘vertical velocity’ directly represents large-scale ascending motion. 

On the other hand, the dominant atmospheric variables also align with the spatial resolution and forecast intervals/periods of the ‘USA’ dataset. Low-resolution, long-term predictions (the ‘USA’ dataset has a spatial resolution of 6 km, with prediction periods of 24 hours) focus on the evolution of large-scale weather patterns. Upper-level variables such as ‘vertical velocity’ can indicate extensive, persistent ascending motion regions, while upper-level ‘u/v wind’ and ‘geopotential’ define the framework of large-scale dynamic processes.

Finally, regarding MRMS QPE and radar data as in Fig. \ref{fig:grad_usa}, the highest attribution is found for near-surface radar observations and MRMS QPE itself. Radar attribution decreases sharply with altitude, which is expected given that the forecast variable--MRMS QPE--is closely linked to near-surface radar observations. Furthermore, as the forecast time interval increases, the attribution magnitude of radar data gradually decreases, whereas that of ERA5 reanalysis data progressively increases. This trend underscores the importance of reanalysis data for long-term precipitation prediction.

These attribution results confirm that the model's reasoning process aligns closely with meteorological principles and underscore the value of incorporating richer atmospheric variables. And they reveal that the latent-space prediction model can adaptively prioritize key atmospheric features according to regional characteristics and task demands. This demonstrates the model's powerful capacity to accurately extract the underlying atmospheric dynamics embedded within the data.

\subsection{Complexity Analyse}

\begin{table}[t]
\vskip 0.01in
\caption{Average computational complexity on ‘USA’ dataset. Para$\rightarrow$Parameters. T-time$\rightarrow$Training time. I-time$\rightarrow$Inference time.}\label{tab1}%
%\vskip 0.15in
\setlength{\tabcolsep}{3pt}
\begin{center}
\begin{small}
% \begin{sc}
\begin{tabular}{cccc}
\toprule
Methods & Para & T-time & I-time \\
 & (M) &  (h) &  (s) \\
\midrule
ConvGRU & \textbf{0.045} & 30.556 & 2.245 \\  % 0812修正
SimVP & 5.827 & \textbf{10.167} & 1.286 \\  % input two time step
SimVP\_HTA & 1.485 & 18.389 & 2.6515 \\  % input one time step
Metnet2 & 13.246  & 20.389 & 5.115 \\
EarthFarseer & 103.576  & 86.667 & 5.318 \\
Ours& 173.148 & 10.944 & \textbf{0.385} \\
\bottomrule
\end{tabular}
% \end{sc}
\end{small}
\end{center}
%\vskip -0.2in
\end{table}
To evaluate the efficiency of our proposed method, we conducted comprehensive comparisons of computational complexity with other approaches, including the number of network parameters, training time, and inference time. As shown in Table \ref{tab1}, although our method contains more parameters than other methods, it demonstrates competitive computational efficiency. Specifically, the training time is only slightly longer than SimVP by 0.777 hours, while the inference time is shorter than all other methods greatly. Such a short inference time can be attributed to the fact that our prediction model operates iteratively in a low-dimensional latent space, whereas most other methods perform iterative predictions in the original high-dimensional physical variable space. More importantly, as shown in the experimental results on two datasets, our method achieves significantly superior prediction performance compared to other approaches. These results collectively demonstrate that our method achieves the optimal prediction accuracy and computational efficiency.

\section{Conclusions}
% 新设计的损失函数的有效性，没有ce loss, 原始ce loss， Focal loss, 新ce loss
% 在隐空间迭代的有效性，隐空间vs原始空间,
% 
% \subsection*{Conclusions and Limitations}\label{sec13}
% 为了解决现有的降水预报模型在短中期预报周期性能差且未关注降水事件服从的极端长尾分布特性的问题，本文加入更多大气观测并构建了在隐空间迭代预测的降水预报模型。相比于原始物理变量空间，隐空间迭代预测，我们的模型可以大大减少模型需要学习的变量演变规律（只学习和降水演变强相关的隐特征的演变规律），加快模型收敛速度。并且针对降水事件的极端长尾分布特性，本研究构建了ELT-CE损失函数来加大对漏检降水事件的惩罚。大量的实验结果充分表明了所提出降水预报模型和损失函数的有效性。
To address the issue that existing precipitation forecasting models perform poorly in short-term forecasting periods, this study incorporates additional atmospheric observations and constructs a precipitation forecasting model that operates through iterative prediction in a latent space. Compared to the original physical variable space, iterative prediction in the latent space allows our model to significantly reduce the number of variable evolution patterns the model needs to learn (focusing only on the evolution patterns of latent features strongly associated with precipitation development), which leads to improved prediction performance and reduced computational overhead. Furthermore, to capture the unbalanced distribution characteristics of precipitation events, this study develops an ‘WMCE’ loss function to increase the penalty for missed precipitation events while providing accurate constraints on precipitation intensity. Extensive experimental results fully demonstrate the superiority and efficiency of the proposed precipitation forecasting model and the loss function. 

Nevertheless, our model still has some aspects that warrant improvement. For instance, the prediction results suffer from overprediction issues, and the spatial resolution of the incorporated atmospheric analysis data (ERA5) is relatively coarse. These areas require further research and enhancement.

% \section*{Declaration statements}
% \section*{Data availability}
% ERA5 data can be downloaded from the ECMWF website:\\ https://cds.climate.copernicus.eu/datasets/reanalysis-era5-pressure-levels?tab=overview. NEXRAD radar data are available via: https://gdex.ucar.edu/datasets/d841000/. MRMS QPE data can be obtained from: https://mtarchive.geol.iastate.edu/. FY-4B satellite imagery is accessible at:\\ https://data.nsmc.org.cn/DataPortal/cn/data/structure.html.

\section{Acknowledgements}
This work was supported by the Fundamental Research Funds for the Central Universities.

% \subsection*{Author Contributions}
% S.L. conducted the main experiments and prepared the manuscript. S.W.L. was responsible for the conceptualization and experimental design. L.L. and W.Z. contributed to data acquisition and processing. J.Y. and M.Z. contributed to the revision of the manuscript.

\section{Competing Interests}
The authors declare that they have no known competing financial interests or personal relationships that could have appeared to influence the work reported in this paper.

{\small
\bibliographystyle{IEEEtran}
\bibliography{test}
}
\end{document}